\documentclass[10pt,twocolumn,letterpaper]{article}

\usepackage{iccv}
\usepackage{times}
\usepackage{epsfig}
\usepackage{graphicx}
\usepackage{amsmath}
\usepackage{amssymb}
\usepackage{booktabs}
\usepackage{adjustbox}
\usepackage{makecell}
\usepackage{multirow}
\usepackage{bbding}
\usepackage{xcolor}
\usepackage{color, colortbl}
\definecolor{citecolor}{HTML}{2980b9}
\definecolor{linkcolor}{HTML}{c0392b}

\usepackage{tabularx}
\usepackage{enumerate}
\usepackage{bbding}
\usepackage{pifont}
\usepackage{wasysym}
\usepackage{float}

\usepackage[hidelinks,pagebackref=true,breaklinks=true,colorlinks,bookmarks=false,citecolor=citecolor,letterpaper=true,linkcolor=linkcolor]{hyperref}

\makeatletter
\newcommand\figcaption{\def\@captype{figure}\caption}
\newcommand\tabcaption{\def\@captype{table}\caption}
\makeatother

\usepackage[capitalize]{cleveref}
\crefname{section}{Sec.}{Secs.}
\Crefname{section}{Section}{Sections}
\Crefname{table}{Table}{Tables}
\crefname{table}{Tab.}{Tabs.}

\iccvfinalcopy 

\def\iccvPaperID{1379} 

\ificcvfinal\fi

\begin{document}

\title{PointCLIP V2: Prompting CLIP and GPT for Powerful\\3D Open-world Learning}

\author{Xiangyang Zhu$^{*1}$, Renrui Zhang$^{*\dagger\ddagger2,3}$, Bowei He$^{1}$, Ziyu Guo$^{2,3}$, Ziyao Zeng$^{5}$, Zipeng Qin$^{2}$\\\vspace{0.2cm} Shanghang Zhang$^{4}$, Peng Gao$^{3}$\\
\normalsize{$^*$ Equal contribution}\quad  $\dagger$ Project leader\quad  $\ddagger$ Corresponding author\vspace{0.3cm}\\
  $^1$City University of Hong Kong \quad \vspace{0.07cm}
  $^2$The Chinese University of Hong Kong\\
  $^3$Shanghai Artificial Intelligence Laboratory \quad
  $^4$Peking University\quad
  $^5$Yale University\vspace{0.2cm}\\
\texttt{\{xiangyzhu6-c, boweihe2-c\}@my.cityu.edu.hk},\\
\texttt{\{zhangrenrui, gaopeng\}@pjlab.org.cn},\quad \texttt{shanghang@pku.edu.cn}
}

\maketitle

\ificcvfinal\thispagestyle{empty}\fi

\begin{abstract}
Large-scale pre-trained models have shown promising open-world performance for both vision and language tasks. However, their transferred capacity on 3D point clouds is still limited and only constrained to the classification task. In this paper, we first collaborate \textbf{CLIP} and \textbf{GPT} to be a unified 3D open-world learner, named as \textbf{PointCLIP V2}, which fully unleashes their potential for zero-shot 3D classification, segmentation, and detection. To better align 3D data with the pre-trained language knowledge, PointCLIP V2 contains two key designs.
For the visual end, we prompt CLIP via a shape projection module to generate more realistic depth maps, narrowing the domain gap between projected point clouds with natural images. For the textual end, we prompt the GPT model to generate 3D-specific text as the input of CLIP's textual encoder. 
Without any training in 3D domains, our approach significantly surpasses PointCLIP by \textbf{+42.90\%}, \textbf{+40.44\%}, and \textbf{+28.75\%} accuracy on three datasets for zero-shot 3D classification.
On top of that, V2 can be extended to few-shot 3D classification, zero-shot 3D part segmentation, and 3D object detection in a simple manner, demonstrating our generalization ability for unified 3D open-world learning. 
Code is available at \url{https://github.com/yangyangyang127/PointCLIP_V2}.
\end{abstract}

\section{Introduction}
\label{sec:intro}

The advancement of spatial sensors has stimulated widespread attention in recent years for both academia and industry. To effectively understand point clouds, the major data form in 3D, many related tasks are put forward and gained great progress, including 3D classification~\cite{qi2016pointnet, wu2019pointconv, zhang2023parameter}, segmentation~\cite{qi2017pointnet++, Xiang2021Walk, wang2019dynamic,wu2022eda}, detection~\cite{xu2020squeezesegv3, meng2021towards}, and self-supervised learning~\cite{zhang2022learning,guo2023joint,zhang2022point,fu2022pos}. Importantly, for the complexity and diversity of open-world circumstances, the collected 3D data normally contains a large number of `unseen' objects, namely, not ever defined and trained by the already deployed 3D systems. Given the human-laboring data annotations, how to recognize such 3D shapes of new categories has become a hot-spot issue, which still remains to be fully explored.

\begin{figure}[t!]
\centering
\includegraphics[width=0.45\textwidth]{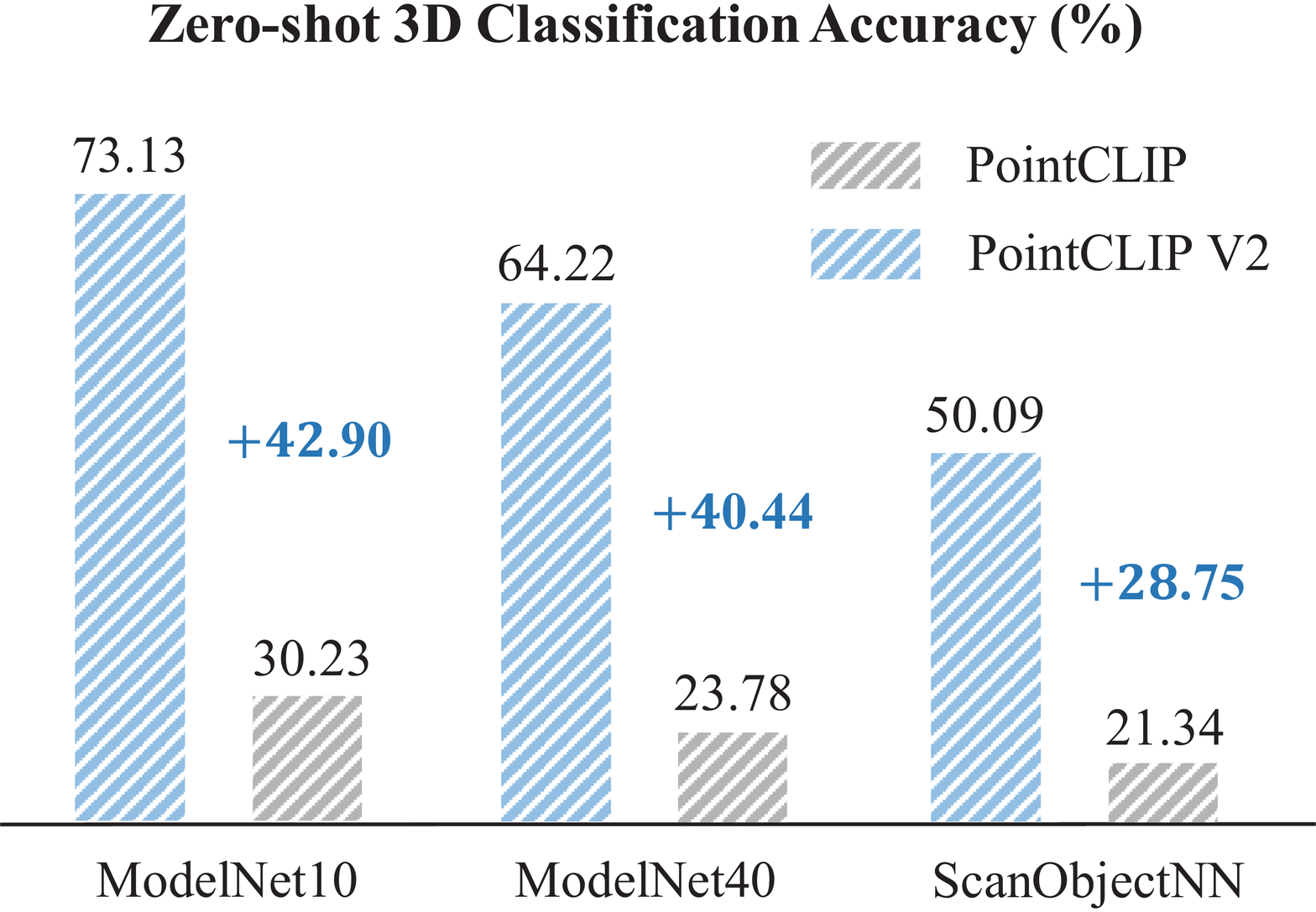}
\vspace{0.1cm}
\caption{\textbf{Zero-shot Performance of PointCLIP V2.} On different 3D datasets, our approach achieves significant accuracy enhancement for zero-shot 3D classification over PointCLIP~\cite{zhang2022pointclip}.}
\label{fig1}
\vspace{-0.2cm}
\end{figure}

Recently, large-scale pre-trained vision and language models, \eg, CLIP~\cite{radford2021learning} and GPT-3~\cite{brown2020language}, have obtained a strong capacity to process data in both modalities. However, limited efforts have focused on their application in the point cloud, and existing work only explores the possibility of CLIP on the 3D classification task, without considering other 3D open-world tasks. PointCLIP~\cite{zhang2022pointclip}, for the first time, indicates that CLIP can be adapted for zero-shot point cloud classification without any 3D training. It projects the `unseen' 3D point cloud sparsely into 2D depth maps, and leverages CLIP's image-text alignment for depth map recognition.  
However, as a preliminary work, the performance of PointCLIP is far from satisfactory as shown in Figure~\ref{fig1}, which cannot be put into actual use. More importantly, PointCLIP only draws support from pre-trained CLIP, without considering the powerful large-scale language model (LLM).
Therefore, we ask the question: \textit{Can we properly unify CLIP and LLM to fully unleash their potentials for unified 3D open-world understanding?}

\begin{figure}[t!]
\centering
\includegraphics[width=0.48\textwidth]{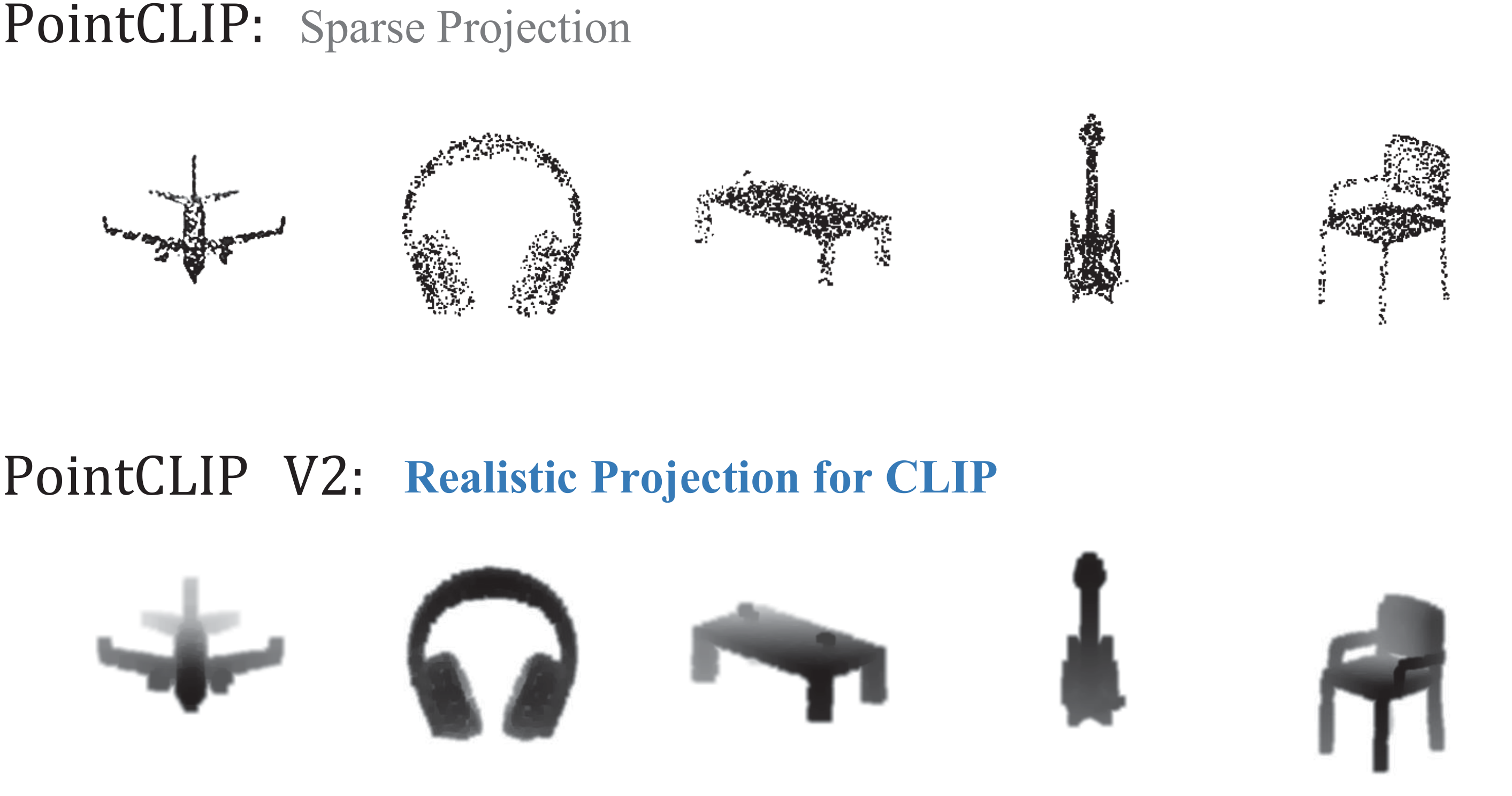}
\caption{\textbf{Comparison of Visual Projection.} PointCLIP V2 (Bottom) generates more realistic depth maps with denser point distribution and smoother depth values.}
\label{fig2}
\vspace{-0.31cm}
\end{figure}

\begin{figure}[t!]
\centering
\includegraphics[width=0.48\textwidth]{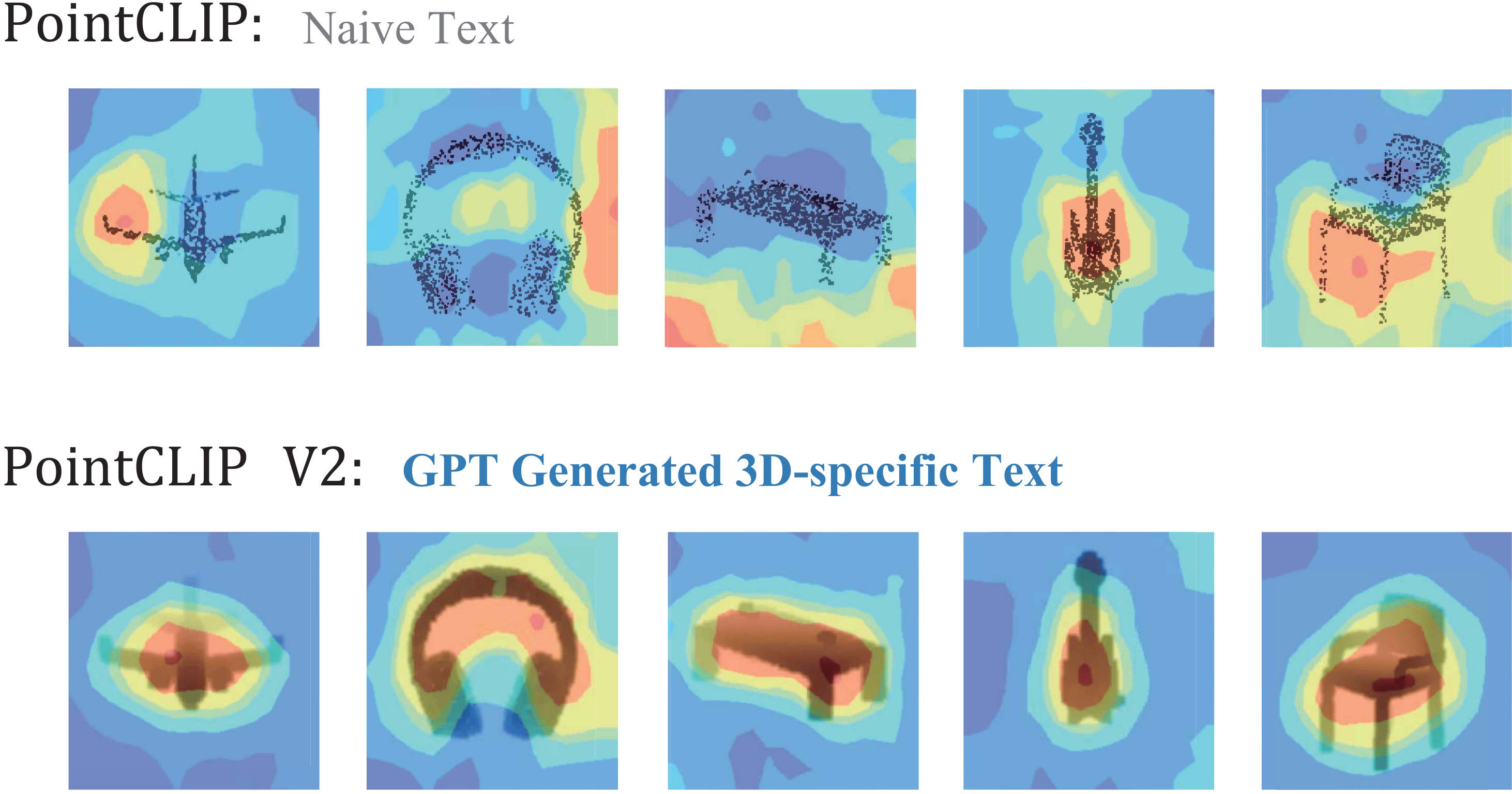}
\caption{\textbf{Comparison of Textual Input.} We visualize the similarity score maps of the encoded textual and visual features, where PointCLIP V2 (Bottom) shows better alignment.}
\vspace{-0.3cm}
\label{fig3}
\end{figure}

We observe that PointCLIP mainly suffers from two factors concerning the 2D-3D domain gap. \textbf{(1) Sparse Projection.} PointCLIP simply projects 3D point clouds onto depth maps as sparsely distributed points with depth values (Figure~\ref{fig2}). Though simple, the scatter-style figures are dramatically different from the real-world pre-training images for both appearances and semantics, which severely confuses CLIP's visual encoder. \textbf{(2) Naive Text.} PointCLIP mostly inherits CLIP's 2D text input, ``\texttt{a photo of a {[CLASS]}.}'' and only appends simple 3D-related words, ``\texttt{a depth map}''. As visualized in Figure~\ref{fig3}, the textual features extracted by CLIP can hardly focus on the target object with high similarity scores. Such naive text cannot fully describe 3D shapes and harms the pre-trained language-image alignment. 

In this paper, we integrate the advantage of CLIP and the GPT-3~\cite{brown2020language} model and propose \textbf{PointCLIP V2}, a powerful framework for unified 3D open-world understanding, including zero-shot/few-shot 3D classification, zero-shot part segmentation, and zero-shot 3D object detection.
Without `seeing' any 3D training data, V2 can project point clouds into realistic 2D figures and align them with 3D-aware text, which fully unleashes CLIP's pre-trained knowledge in the 3D domain.

Firstly, we propose to \textbf{Prompt CLIP with Realistic Projection}, which generates CLIP-preferred images from 3D point clouds. Specifically, we transform the irregular point cloud into grid-based voxels and then apply non-parametric 3D local filtering on top. By this, the projected 3D shapes are composed of denser points with smoother depth values. As shown in Figure~\ref{fig2}, our generated figures are more visually similar to real-world images and can highly unleash the representation capacity of CLIP's pre-trained visual encoder.
Secondly, we \textbf{Prompt GPT with 3D Command} to generate text with rich 3D semantics as the input of CLIP's textual encoder. By feeding heuristic 3D-oriented command into GPT-3, \eg, ``\texttt{Give a caption of a table depth map:}'', we leverage its language-generative knowledge to obtain a series of 3D-specific text, \eg, ``\texttt{A height map of a table with a top and several legs.}''. A group of language commands is customized to prompt GPT-3 to produce diverse text with 3D shape information.
As shown in Figure~\ref{fig3}, the textual features of PointCLIP V2 exert stronger matching properties to the projected maps, largely boosting CLIP's image-text alignment for 3D point clouds.

With our prompting schemes, PointCLIP V2 exhibits superior performance for zero-shot 3D classification, surpassing PointCLIP by $+42.90\%$, $+40.44\%$, and $+28.75\%$ accuracy, respectively on ModelNet10~\cite{wu20153d}, ModelNet40~\cite{wu20153d}, and ScanObjectNN~\cite{uy2019revisiting} datasets. Further, our approach can be adapted for more no-trivial 3D open-world tasks by marginal modifications, such as a learnable 3D smoothing for 3D few-shot classification, a back-projection head for zero-shot segmentation, and a 3D region proposal network for zero-shot detection. This fully indicates the power of V2 for general 3D open-world understanding.

Our contributions are summarized as follows:

\begin{itemize}
    \item We propose PointCLIP V2, a powerful cross-modal learner unifying CLIP and GPT-3 to transfer the pre-trained vision-language knowledge into 3D domains. 
    
    \item We introduce a realistic projection to prompt CLIP and 3D-oriented command to prompt GPT-3 to effectively mitigate the domain gap among 2D, 3D, and language.
    
    \item As the first work for unified 3D open-world learning, our PointCLIP V2 can be further extended for zero-shot part segmentation and 3D object detection.
\end{itemize}

\section{Related Works}
\label{sec:related work}

\begin{figure}[t!]
\centering
\includegraphics[width=0.462\textwidth]{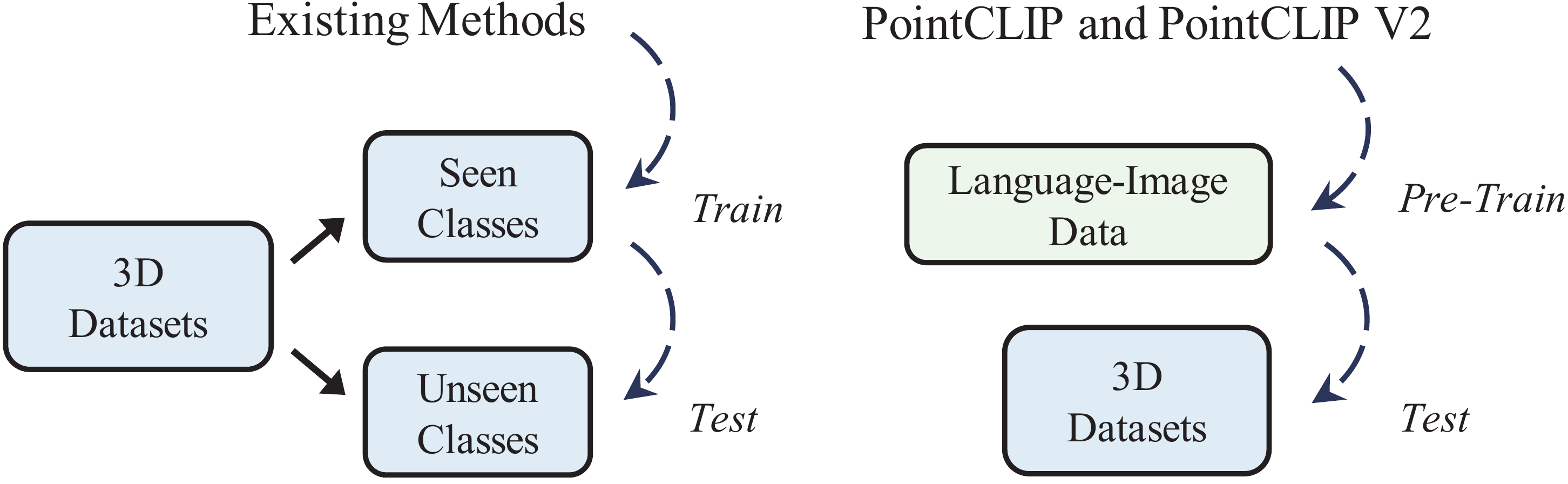}
\vspace{0.07cm}
\caption{\textbf{Comparison of Open-world Settings.}
Existing methods still depend on prerequisite 3D training to recognize the `unseen' point clouds. In contrast, we require no training in the 3D domain and directly conduct 3D open-world understanding.}
\label{fig:scheme_comparison}
\vspace{-0.22cm}
\end{figure}

\paragraph{3D Open-world Learning.}
Traditional methods for 3D open-world learning still require 3D training data as a pre-training stage. The series of work of Cheraghian \etal train zero-shot classifiers on `seen' 3D categories by maximizing inter-class divergence in latent space, and then test on `unseen' ones \cite{cheraghian2019mitigating, cheraghian2019zero, cheraghian2022zero}. Some recent works \cite{michele2021generative,liu2021segmenting,naeem20223d, Lu2022Open} also investigate open-world semantic segmentation and 3D object detection for more complex 3D scenes.
Inspired by CLIP-based adaption methods~\cite{zhang2021tip,gao2021clip,lin2022frozen}, PointCLIP \cite{zhang2022pointclip} achieves zero-shot point cloud recognition without any training on 3D datasets. By transferring the pre-trained CLIP model~\cite{radford2021learning}, the 2D knowledge can be effectively utilized for recognizing 3D data. CLIP2Point \cite{Huang2022CLIP} further improves the adaption performance of CLIP on point clouds by an additional 3D pre-training.
In this paper, we propose PointCLIP V2 and follow the open-world setting of PointCLIP, which is more challenging than previous methods as compared in Figure~\ref{fig:scheme_comparison}.
We require no `seen' 3D training and, for the first time, simultaneously conduct zero-shot 3D part segmentation and object detection, achieving unified 3D open-world understanding.

\paragraph{Projection for Point Clouds.}
Concurrent to point-based methods~\cite{qi2016pointnet, qi2017pointnet++,  ma2022rethinking}, projection-based point cloud analysis aims to utilize plentiful 2D networks for 3D domains by projecting point clouds into 2D images \cite{su2015multi, Shi2015DeepPano, yang2019learning, wang2019dominant, wei2020view, ahmed2019epn}. 
Therein, PointCLIP~\cite{zhang2022pointclip} follows SimpleView~\cite{goyal2021revisiting} to conduct perspective transformation as 3D-to-2D projection, which achieves high efficiency but limited classification accuracy. 
Under 3D open-world settings, we are motivated to develop more efficient and realistic projection methods for prompting CLIP on point cloud data. In Table~\ref{table:ablation_projection}, we compare our approach with existing advanced projection methods for latency and accuracy. For a fair comparison, we implement all prior works under the pipeline of our V2, namely, with our GPT prompting approach to fully reveal their effectiveness. As shown, our realistic projection exhibits faster inference speed than other approaches and attains higher zero-shot performance than PointCLIP.

\paragraph{Prompt Learning in Vision.}
Prompt engineering first derives from natural language processing, where a textual template, termed as prompt, is generated to narrow the domain gap between the pre-training pre-text task and downstream scenarios~\cite{liu2021pre,Jiang2020How,wallace2019universal,Jiang2020How}. Inspired by this, CoOp~\cite{zhou2022coop} firstly introduces learnable prompting into 2D vision-language classification, and the follow-up CoCoOp~\cite{zhou2022cocoop} extends it for 2D domain generalization. CuPL~\cite{pratt2022Whatdoes} and CaFo~\cite{zhang2023prompt} leverage GPT-3~\cite{brown2020language} to enhance the downstream performance of CLIP on various 2D datasets. From another perspective, visual prompting methods propose to append input images with learnable visual pixels~\cite{jia2022visual, bahng2022visual, chen2022visual, gan2023decorate} or embeddings~\cite{jia2022visual,guo2023viewrefer,zhang2023personalize}, and improve pre-trained vision backbones without downstream fine-tuning. In this paper, we seek to prompt both CLIP's visual encoder by realistic projection and textual encoder by GPT-3 to improve its zero-shot prediction. 

\paragraph{GPT-3 Model.} The Generative Pre-trained Transformer (GPT) models~\cite{radford2018improving, radford2019language, brown2020language} have achieved a progressive improvement in processing natural languages. Among them, GPT-3 demonstrates a remarkable proficiency in both language comprehension and generation, compared to its predecessors~\cite{radford2018improving, radford2019language, liu2019roberta, yang2019xlnet, raffel2020exploring}. GPT-3 is a large-scale autoregressive language model of 175 billion trainable parameters. Although not open-sourced, some efforts have explored its application to downstream tasks, such as PICa~\cite{yang2022empirical} for visual question-answering, CuPL~\cite{pratt2022Whatdoes} for 2D zero-shot recognition, and CaFo~\cite{zhang2023prompt} for 2D few-shot learning. In this work, we, for the first time, prompt GPT-3~\cite{brown2020language} to boost open-world 3D tasks via 3D-related command.

\begin{table}[t!]
\centering
\begin{adjustbox}{width=0.9\linewidth}
	\begin{tabular}{lccc}
	\toprule
		\makecell*[c]{Method} & Latency & ModelNet40 & ScanObjectNN \\ \cmidrule(lr){1-1} \cmidrule(lr){2-2} \cmidrule(lr){3-4}
		Phong Shading~\cite{su2015multi} & $107.2$ & $57.30$ & $29.33$\\
		Height Map~\cite{su2018adeeper} & $87.7$ & $54.73$ & $26.25$\\
		Silhouette Map~\cite{su2018adeeper} & $87.9$ & $48.40$ & $20.91$ \\
		PointCLIP~\cite{zhang2022pointclip} & $\mathbf{11.3}$ & $42.53$ & $26.37$\\
		\textbf{PointCLIP V2} & $16.7$ & $\mathbf{64.22}$ & $\mathbf{35.36}$\\
	\bottomrule
	\end{tabular}
\end{adjustbox}
\vspace{0.23cm}
\caption{\textbf{Comparison of Different Projection Methods.}
We report zero-shot classification results (\%) on two datasets~\cite{wu20153d, uy2019revisiting}, and compare the inference latency (ms) by projecting 10-view images from an input point cloud.}
\label{table:ablation_projection}
\vspace{-0.2cm}
\end{table}

\section{Methods}
\label{sec:methods}
The overall framework of PointCLIP V2 is shown in Figure \ref{fig:framework2}. Inheriting from CLIP~\cite{radford2021learning}, our framework consists of two pre-trained visual and textual encoders. To bridge the modal gap, we introduce a realistic projection (Sec.~\ref{method:shape_projection}) from 3D to depth maps, and GPT-generated 3D-specific text (Sec.~\ref{method:3D_prompt}) to align depth maps with languages. PointCLIP V2 can also be extended to various 3D tasks for unified 3D open-world learning (Sec.~\ref{s3.4}).

\subsection{Prompting CLIP with Realistic Projection}
\label{method:shape_projection}

\begin{figure}[t!]
\centering
\includegraphics[width=0.399\textwidth]{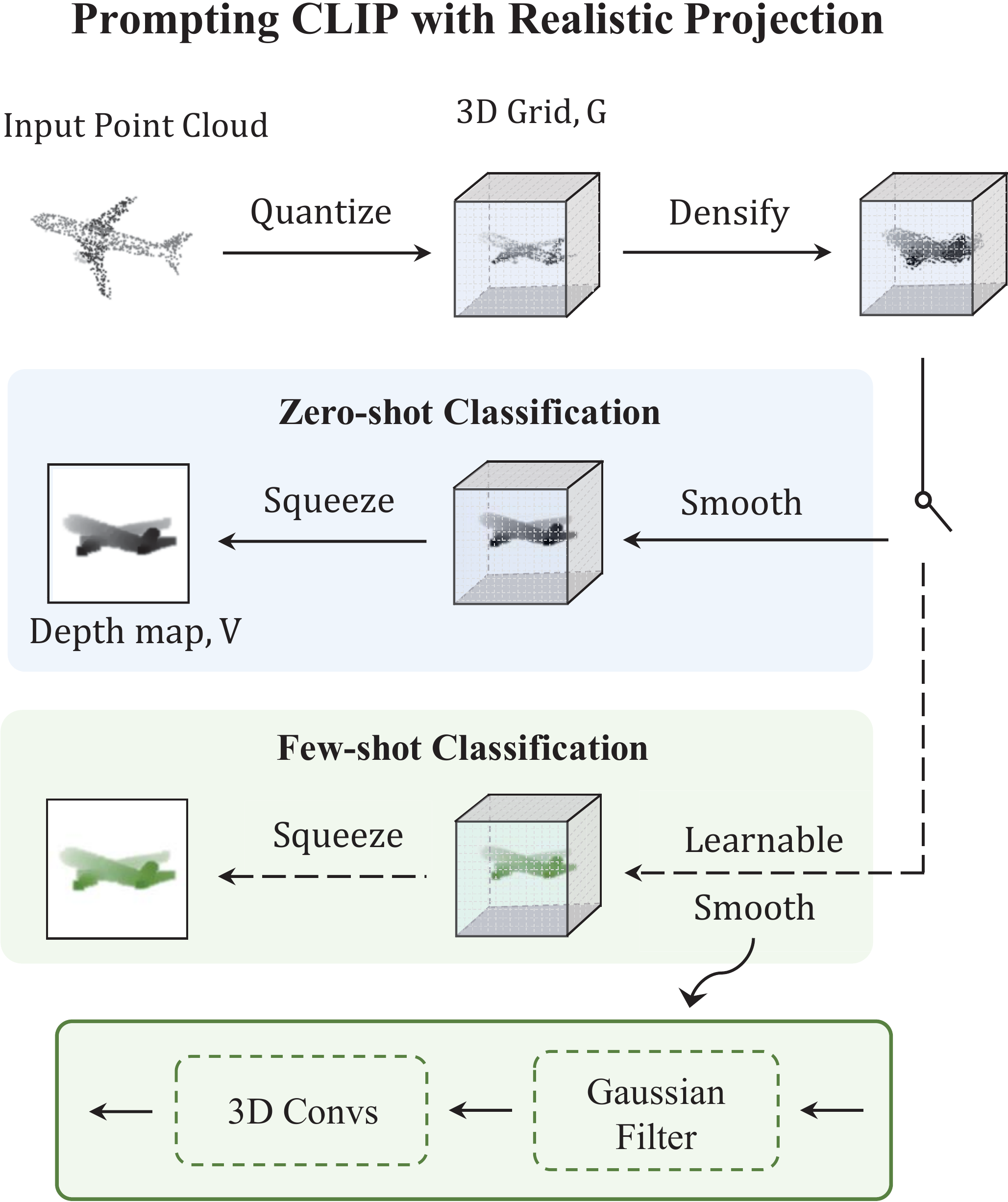}
\vspace{0.19cm}
\caption{\textbf{Prompting CLIP with Realistic Projection.} We present the projection pipeline for one of the views. The switch selects zero- or few-shot classification with learnable smoothing.}
\label{fig:shape_projection}
\vspace{-0.1cm}
\end{figure}

To generate more realistic 2D input from 3D data for CLIP and also achieve time efficiency, we project 3D point clouds into depth maps by four steps: Quantize, Densify, Smooth, and Squeeze, as shown in Figure \ref{fig:shape_projection}.

\vspace{-6pt}
\paragraph{Quantize.} 
For different $M$ views to be projected, we respectively create a zero-initialized 3D grid $G \in \mathbb{R}^{H \times W \times D}$, where $H, W, D$ denote its spatial resolutions and $D$ specially represents the depth dimension vertical to the view plane.
Taking one view as an example, we normalize the 3D coordinates of the input point cloud into $[0,1]$ and project a point $p=(x,y,z)$ into a voxel in the grid by
\begin{align}
G(\left \lceil sHx \right \rceil, \left \lceil sWy \right \rceil, \left \lceil Dz \right \rceil)=z,
\label{eq:quantize}
\end{align}
where the voxels are assigned with different depth values, and $s \in (0,1]$ denotes a scale factor to adjust the projected shape size. For multiple points projected into the same voxel, we simply assign the minimum depth value. This is because, from the perspective of the target image plane, the points with a smaller depth value $z$ would occlude the larger ones.
Then, we obtain a 3D grid $G$ containing sparse depth values, most voxels of which are empty due to the sparsity of point clouds.

\vspace{-6pt}
\paragraph{Densify.} 
To tackle such unreal scattering, we densify the grid via a local mini-value pooling operation to guarantee visual continuity. We reassign every voxel in $G$ by the minimum voxel value within a local spatial window. Likewise, compared to the average and max pooling, preserving the minimum depth values accords with the occluded visual appearances on the projected maps. In this way, the originally vacant voxels between the sparse points can be effectively filled with reasonable depth values, while the background voxels still remain empty, which derives dense and solid spatial shape representations.

\label{method:tasks}
\begin{figure}[t!]
\centering
\includegraphics[width=0.409\textwidth]{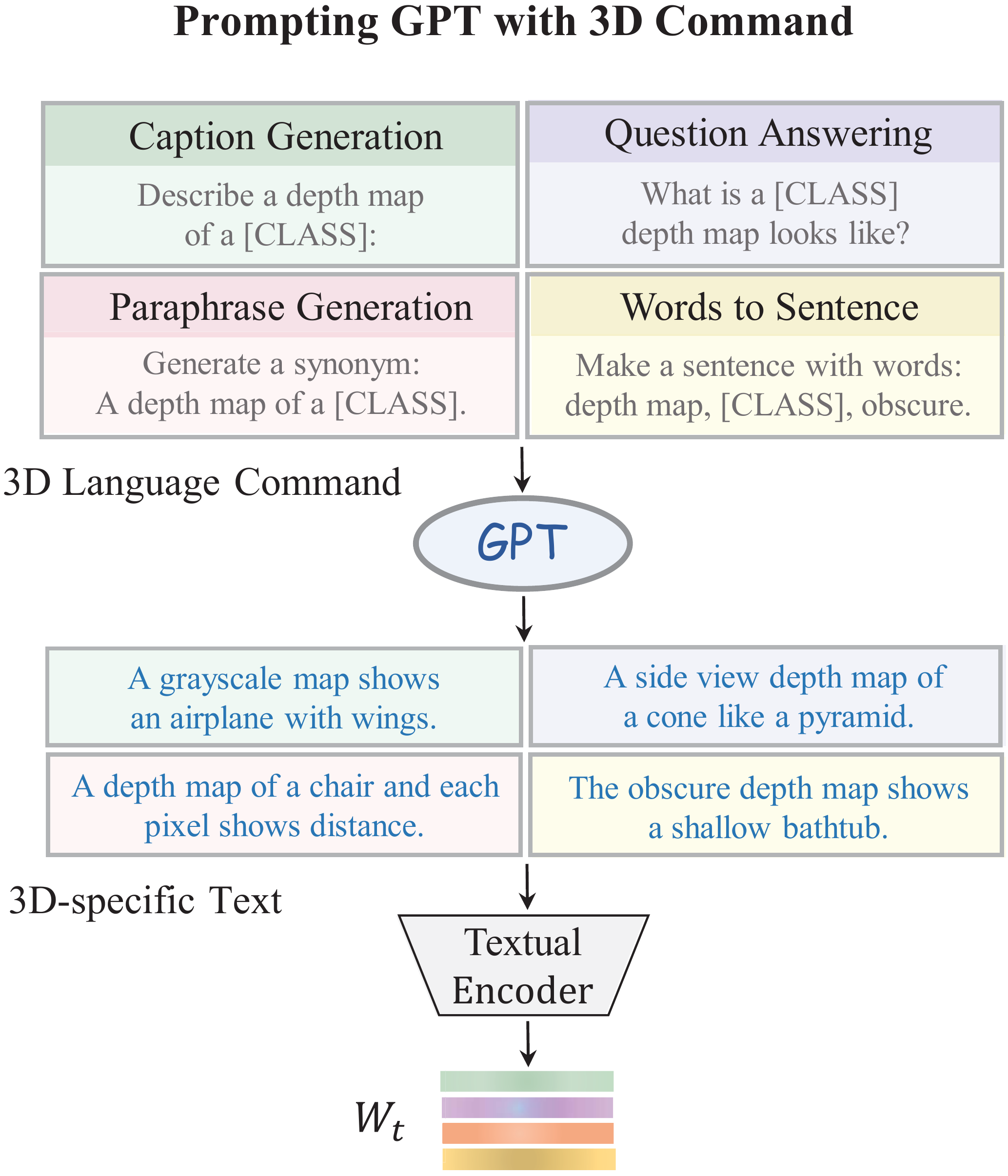}
\vspace{0.15cm}
\caption{\textbf{Prompting GPT with 3D Command.} We feed four types of language command into the pre-trained GPT-3, which generates a series of 3D-specific text for CLIP's textual encoder.}
\label{fig:3d_command}
\vspace{-0.25cm}
\end{figure}

\vspace{-6pt}
\paragraph{Smooth.} 
As the local pooling operation might introduce artifacts on some 3D surfaces, we adopt a non-parametric Gaussian kernel for shape smoothing and noise filtering. With a proper kernel size and variance, the filtering can not only remove the spatial noises caused by densification but also preserve the sharpness of edges and corners in the original 3D shapes. By this, we acquire a more compact and smooth shape represented by the 3D grid.

\begin{figure*}[t!]
\centering
\includegraphics[width=0.99\textwidth]{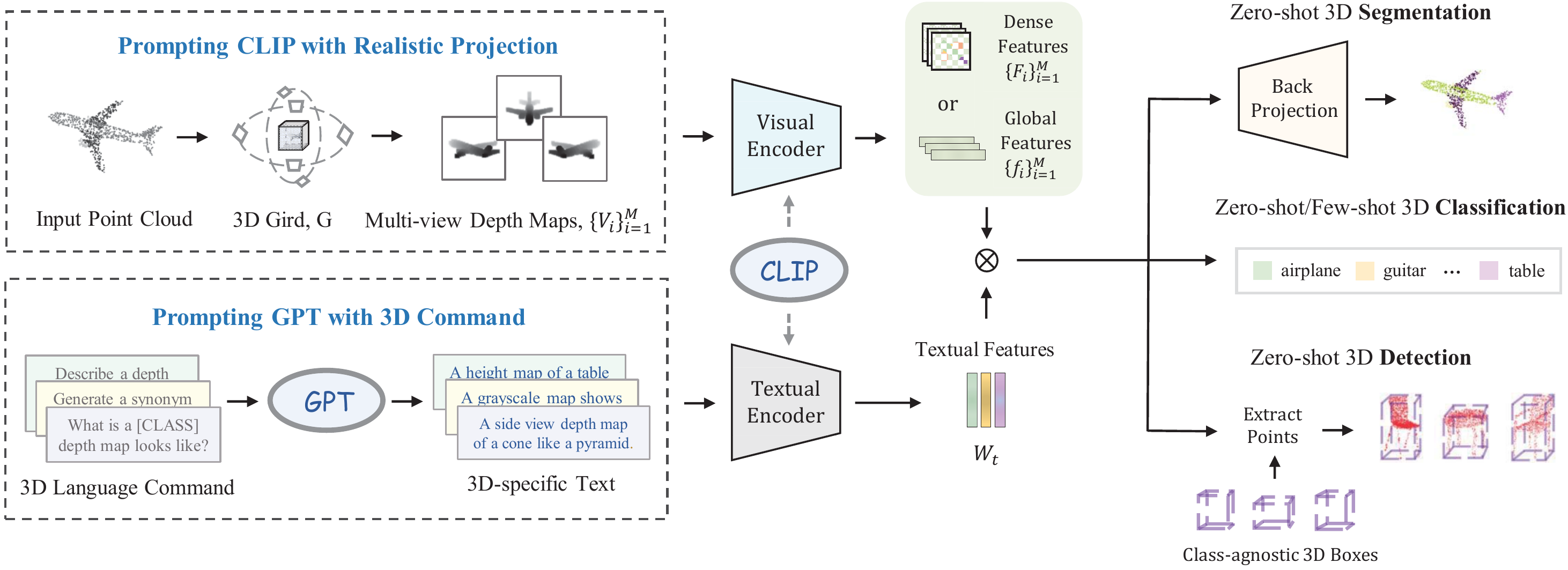}
\vspace{0.1cm}
\caption{\textbf{The Unified Framework of PointCLIP V2 for 3D Open-world Learning.}
We first generate high-quality depth maps via a realistic projection to prompt CLIP's~\cite{radford2021learning} visual encoder. Then, we design 3D language command to prompt GPT-3~\cite{brown2020language} for 3D-specific text into CLIP's textual encoder. V2 can also be extended to 3D segmentation and detection by simple modifications.}
\label{fig:framework2}
\vspace{-0.1cm}
\end{figure*}

\vspace{-6pt}
\paragraph{Squeeze.} 
As the final step, we simply squeeze the depth dimension of ${G}$ to acquire the projected depth map $V \in \mathbb{R}^{H \times W}$. We extract the minima of every depth channel as the value for each pixel location and repeat it for three times as the RGB intensity. Our grid-based projection can be simply achieved by a minimum pooling along the depth channel of $G$, more friendly for hardware implementation.

\subsection{Prompting GPT with 3D Command}
\label{method:3D_prompt}

To better activate CLIP's textual encoder to align with our depth maps, we aim to utilize 3D-specific description with category-wise shape characteristics as the textual input of CLIP, instead of using the general ``\texttt{a photo of a [CLASS]:}''. Considering the powerful descriptive capacity of LLMs, we leverage GPT-3~\cite{brown2020language} to generate 3D-specific text with sufficient 3D semantics for CLIP's textual encoder as shown in Figure~\ref{fig:3d_command}. Normally, GPT-3 receives a language command and outputs a response via pre-trained knowledge. To fully adapt GPT-3 to 3D domains, we propose the following four series of heuristic command:

\vspace{-6pt}
\paragraph{Caption Generation.} 
Given a descriptive command, GPT-3 synthesizes general captions for the target projected 3D shape, \eg, Input: ``\texttt{Describe a depth map of a {[window]}:}''; GPT-3: ``\texttt{It depicts the {[window]} as a dark pane.}''.

\vspace{-6pt}
\paragraph{Question Answering.} 
GPT-3 produces descriptive answers to the 3D-related question, \eg, Input: ``\texttt{How to describe a depth map of a [table]}?''; GPT-3: ``\texttt{The {[table]} may have a rectangular or circular flat top and legs.}''.

\vspace{-6pt}
\paragraph{Paraphrase Generation.} 
For a depth map description, GPT-3 is expected to generate a synonymous sentence. \eg, Input: ``\texttt{Generate a synonym for the sentence: A grayscale depth map of an inclined {[bed]}.}''; GPT-3: ``\texttt{An monochrome depth map of an oblique {[bed]}.}''.

\vspace{-6pt}
\paragraph{Words to Sentence.} 
Based on a group of keywords, GPT-3 is requested to organize them into a complete sentence and enrich additional shape-related contents, \eg, Input: ``\texttt{Make a sentence using these words: a {[table]}, depth map, smooth.}''; GPT-3: ``\texttt{This smooth depth map shows a {[table]} at the corner.}''. The adjective ``\texttt{smooth}'' here depicts the natural appearance caused by the smoothing operation.

\vspace{0.45cm}
For a $K$-category 3D dataset, we place $K$ category names at the ``\texttt{{[CLASS]}}'' position of each command and feed them into GPT-3, which generates 3D-specific descriptions with rich category-wise semantics. Finally, we integrate the descriptions of each category and regard them as the input for CLIP's textual encoder.

\begin{table*}[ht!]
\centering
\begin{adjustbox}{width=1.0\linewidth}
	\begin{tabular}{lccccccc}
	\toprule
		Method & 2D Pre-train & 3D Pre-train &ModelNet10 & ModelNet40 &S-OBJ\_ONLY &S-OBJ\_BG &S-PB\_T50\_RS \\
		\cmidrule(lr){1-1}
		\cmidrule(lr){2-3} 
		\cmidrule(lr){4-8} 
	    
	    CLIP2Point~\cite{Huang2022CLIP} &\checkmark &\checkmark & $66.63$ & $49.38$ & $35.46$ &$30.46$ & $23.32$ \\
	    Cheraghian~\cite{cheraghian2022zero} & - & \checkmark &$68.50$ & - &- &- &-  \\
     \cmidrule(lr){1-8}
     PointCLIP \cite{zhang2022pointclip}\vspace{0.05cm} &\checkmark & - &$30.23$ & $23.78$  &$21.34$ &$19.28$ & $15.38$ \\
	    \textbf{PointCLIP V2}\vspace{0.1cm} &\checkmark & - &$\mathbf{73.13}$ &$\mathbf{64.22}$  &$\mathbf{50.09}$ &$\mathbf{41.22}$ &$\mathbf{35.36}$\\
     \textit{Improvement} &&&\textcolor{blue}{$+42.90$}&\textcolor{blue}{$+40.44$}&\textcolor{blue}{$+28.75$}&\textcolor{blue}{$+21.94$}&\textcolor{blue}{$+19.98$}\\
	\bottomrule
	\end{tabular}
\end{adjustbox}
\vspace{0.1cm}
\caption{\textbf{Zero-shot 3D Classification (\%) ModelNet10~\cite{wu20153d}, ModelNet40~\cite{wu20153d} and ScanObjectNN~\cite{uy2019revisiting}}. We report the performance of other methods with their \textbf{\textit{best-performing settings, \eg, visual encoder, projected view number, and textual input}}. ``2D Pre-train'' denotes the pre-training of CLIP on image-language pairs, and ``3D Pre-train'' denotes the training on 3D datasets.}
\label{table:zero_shot_classification}
\vspace{-0.1cm}
\end{table*}

\subsection{Unified Open-world Learning}
\label{s3.4}

By introducing the realistic projection and 3D-specific text, PointCLIP V2 exhibits strong generalization capacity and can be adapted for different 3D open-world tasks.

\paragraph{3D Zero-shot Classification.}
For all $M$ views in the visual branch, we feed the projected depth maps $\left\{V_{i} \right \}_{i=1}^{M}$ into CLIP's visual encoder and obtain the multi-view features $\left \{f_{i} \right \}_{i=1}^{M}$, where $f_{i} \in \mathbb{R}^{1 \times C}$.
For the textual branch, we leverage CLIP's textual encoder to extract the category feature $W_{t} \in \mathbb{R}^{K \times C}$, which serves as the zero-shot classification weights.
Then, the final zero-shot classification $\mathrm{logits}$ are calculated by aggregating the multi-view alignment between $\left \{f_{i} \right \}_{i=1}^{M}$ and $W_{t}$, formulated as
\begin{align}
\label{equa:zeroshot}
\begin{split}
    &\mathrm{logits} = \sum\nolimits_{i=1}^M{\alpha_i\cdot f_i W_t^T} \in \mathbb{R}^{1 \times K},
\end{split}
\end{align}
where $\alpha_i$ denotes a hyper-parameter weighing the importance of view $i$. 

\paragraph{3D Few-shot Classification.}
Given a small set of 3D training data, we can modify our smoothing operation of the realistic shape projection to be learnable, as shown in Figure~\ref{fig:shape_projection}. Specifically, as the irregular point clouds have been converted into grid-based voxels, we adopt two 3D convolutional layers after the Gaussian filter. Such learnable modules can summarize the 3D geometric knowledge from the few-shot dataset, and further adapt the 3D shape to be more CLIP-friendly. During training, we freeze the two encoders of CLIP to preserve the pre-trained knowledge and avoid over-fitting on small-scale few-shot data.

\paragraph{3D Zero-shot Part Segmentation.}
Besides shape classification, we propose a zero-shot segmentation pipeline for our framework, which can also work for the existing PointCLIP. Instead of the global features $\left \{f_{i} \right \}_{i=1}^{M}$, we adopt CLIP's visual encoder to extract dense features $\left \{F_{i} \right \}_{i=1}^{M}$ from $\left\{V_{i} \right \}_{i=1}^{M}$, where $F_i \in \mathbb{R}^{H \times W \times C}$. Specifically, we output the feature maps from the visual encoder before its final pooling operation and upsample the features into the original depth map size. 
For our 3D-specific text, we utilize GPT-3 to generate the descriptions for different part categories. As an example, for a part category ``\texttt{[PART]}'' within object ``\texttt{[CLASS]}'', we construct the command as ``\texttt{Describe the {[PART]} part of a {[CLASS]} in a depth map:}''.
Then, for view $i$, we conduct dense alignment between each pixel and the textual feature $W_{t}$, \ie, segmenting different parts of the shape on multi-view depth maps, formulated as
\begin{align}
\begin{split}
    \mathrm{logits}_{i} = F_i W_t^T \in \mathbb{R}^{H \times W \times K}.
\end{split}
\end{align}
Each element in $\mathrm{logits}_{i}$ denotes the pixel-wise classification logits. After this, we back-project the logits of different views into the 3D space according to the 2D-3D correspondence. As one view can only depict a partial point cloud due to occlusion, we average the prediction across different views for each point, where we acquire the final part segmentation logits in 3D space. Via the geometric back projection, the segmentation task in 3D can be tackled in a zero-shot manner.

\vspace{-0.2cm}
\paragraph{Zero-shot 3D Object Detection.}
For 3D object detection, we follow the settings of 2D open-world detection~\cite{gu2021open, zhong2022regionclip} to equip our V2 as a zero-shot classification head on top of pre-trained region proposal networks (RPN). 
We first utilize 3DETR~\cite{misra2021end} as the 3D RPN to generate class-agnostic 3D box candidates.
Then, we extract the raw points within each 3D box and feed them into V2 for zero-shot classification. 
By this, the V2-based 3DETR can detect `unseen' objects in a zero-shot manner.

\section{Experiments}
\label{sec:experiments}

In this section, we first illustrate the detailed network configurations of PointCLIP V2, and then present our open-world performance on different 3D tasks.

\subsection{Implementation Details}
\paragraph{CLIP Prompting.} 
We follow PointCLIP~\cite{zhang2022pointclip} to project a point cloud into depth maps of $10$ views. We set the size of grid $G$ as $H \times W \times D = 112 \times 112 \times 8$, and the projected depth map is upsampled to $224 \times 224$. The point cloud is placed at the center of the grid, and the scale factor $s$ is set to $0.7$ for better visual appearances. The window size of the minimum pooling for densifying is $(6, 6, 2)$. The kernel size of the Gaussian filter is set as $(3, 3, 1)$. We randomly sample $1024$ points as input and adopt Vision Transformer \cite{dosovitskiy2020image} with patch size $16\times 16$ as default, denoted as ViT-B/16. 

\vspace{-0.1cm}
\paragraph{GPT Prompting.} 
We design 50 different 3D language commands, containing $13$ for caption generation, $13$ for question answering, $12$ for paraphrase generation, and $12$ for words-to-sentence. Each command triggers GPT-3 to produce $20$ 3D-specific descriptions, and we finally obtain around $250$ descriptions for each command type and $1000$ descriptions in total for one category. We use ``text-davinci-002'' GPT-3 engine and set the temperature constant to $0.7$. The largest length of a 3D-specific description is set to $40$. For the textual encoder, a 12-layer transformer~\cite{vaswani2017attention} is adopted to encode our generated text.

\begin{figure*}[t]
\vspace{5pt}
\begin{minipage}[c]{0.60\textwidth}
\centering
\hspace{-10pt}
\includegraphics[width=4.9cm]{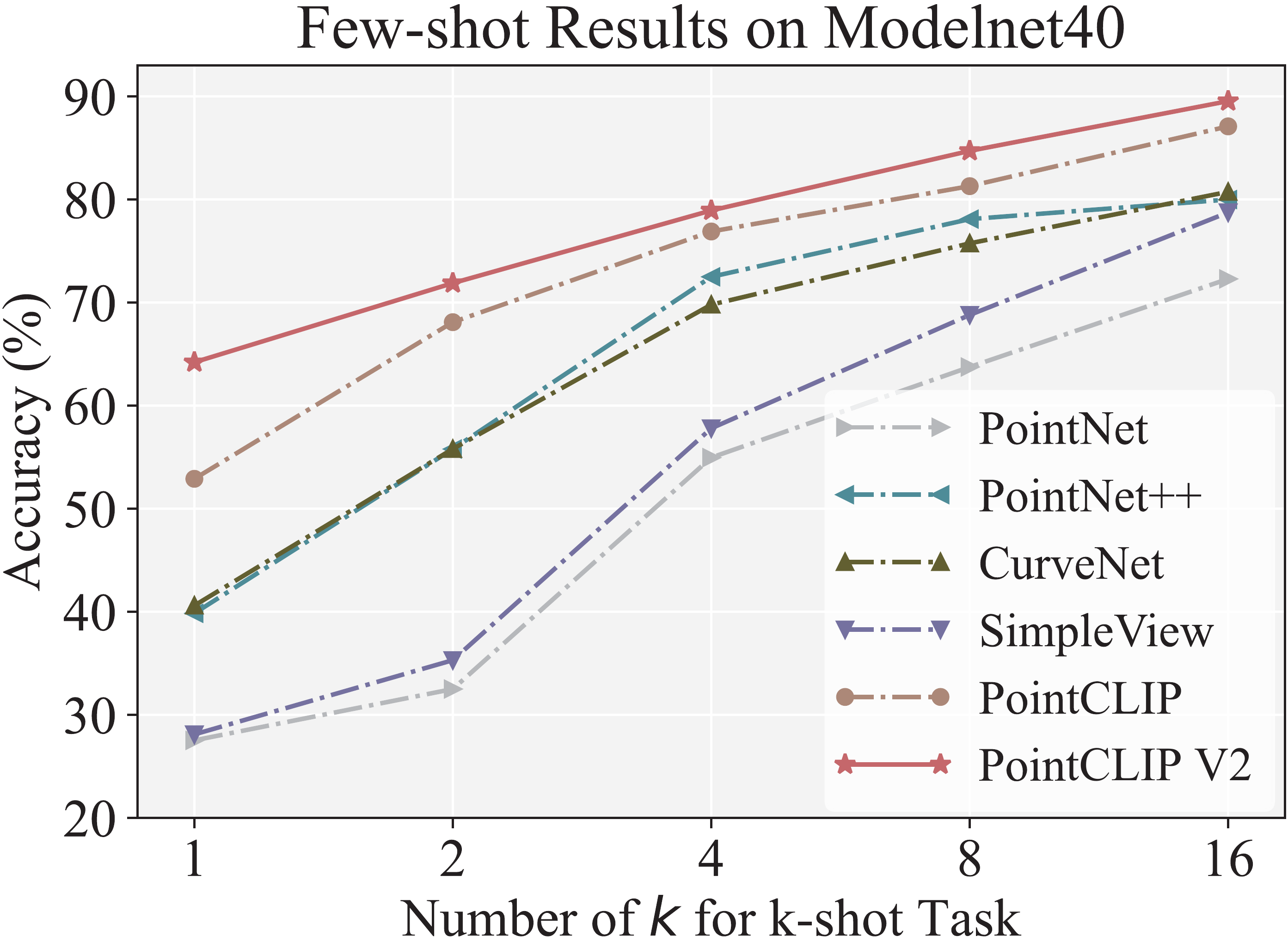}\hspace{10pt}\includegraphics[width=4.9cm]{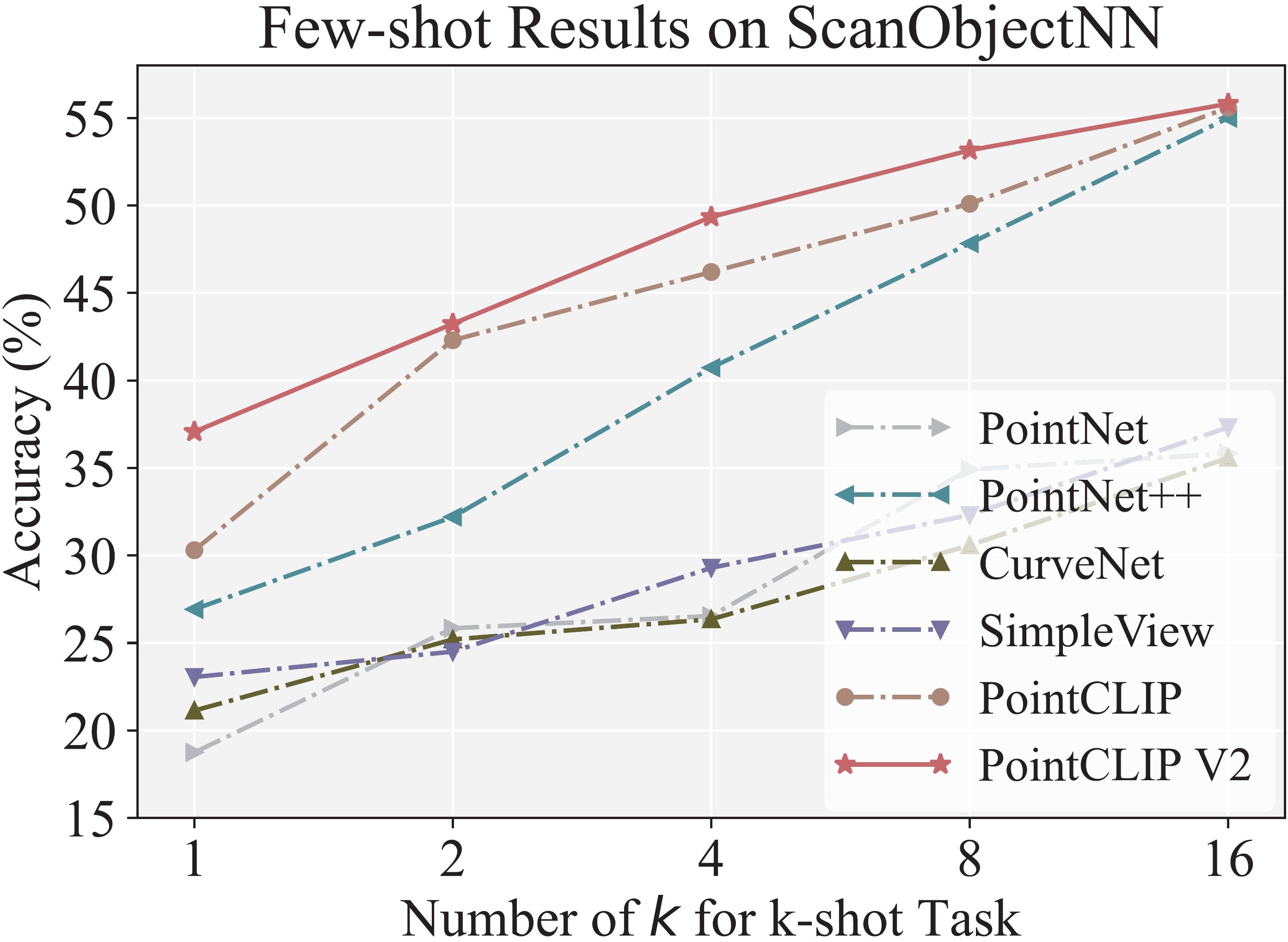}
\caption{\textbf{Few-shot 3D Classification on ModelNet40~\cite{wu20153d} and ScanObjectNN~\cite{uy2019revisiting}.} We adopt the PB\_T50\_RS split of ScanObjectNN for comparison.}
\label{fig:few_shot_classification}
\end{minipage}\hspace{10pt}
\begin{minipage}[c]{0.33\textwidth}
\vspace{9pt}
\begin{adjustbox}{width=5.9cm}
	\begin{tabular}{cccc}
	\toprule
	\makecell[c]{Learnable\\ Smooth} & \makecell[c]{View\\ Weighing} & \makecell[c]{GPT\\ Prompting} & \makecell[c]{$16$-shot} \\ \cmidrule(lr){1-4}
	- & - & - & $85.52$\\
	\checkmark & - & - & $86.22$\\
	\checkmark & \checkmark & - & $87.11$\\
	\checkmark & - & \checkmark & $89.55$\\ 
	\checkmark & \checkmark & \checkmark & $89.55$ \\ 
	\bottomrule
	\end{tabular}
	\end{adjustbox}
	\vspace{5pt}
    \tabcaption{\textbf{Ablation Study of Few-shot Learning on ModelNet40~\cite{wu20153d}.} We report the $16$-shot classification accuracy (\%). }
\label{table:few_ablation}
\end{minipage}\hspace{17pt}
\end{figure*}

\subsection{Zero-shot Classification}
\label{zero-exp_sec}

\begin{table}[t!]
\centering
\begin{adjustbox}{width=0.89\linewidth}
	\begin{tabular}{ccccc}
	\toprule
		Quantize & Densify & Smooth & Squeeze & Zero-shot \\ \cmidrule(lr){1-5}
		- & Min & \checkmark & - & $57.35$\\
		\checkmark & - & - & \checkmark & $44.50$\\
		\checkmark & Min & - & \checkmark & $59.64$ \\
		\checkmark & - & \checkmark & \checkmark & $50.20$\\
		\checkmark & Max & \checkmark & \checkmark & $57.35$ \\
		\checkmark & Avg & \checkmark & \checkmark & $60.71$ \\
  \checkmark & Min & \checkmark & \checkmark & $64.22$ \\
	\bottomrule
	\end{tabular}
\end{adjustbox}
\vspace{0.2cm}
\caption{\textbf{Ablation Study of Realistic Shape Projection} on ModelNet40~\cite{wu20153d} zero-shot classification (\%). We compare the four steps in our projection module.}
\vspace*{-5pt}
\label{table:shape_projection}
\end{table}

\paragraph{Settings.}
The zero-shot classification performance is evaluated on three widely-used benchmarks: ModelNet10~\cite{wu20153d}, ModelNet40~\cite{wu20153d} and ScanObjectNN~\cite{uy2019revisiting}. Three splits of the ScanObjectNN dataset are investigated: OBJ\_ONLY, OBJ\_BG, and PB\_T50\_RS. Following the zero-shot principle, we directly test the classification performance on the full test set without learning from the training set. We compare existing methods under their best settings. Specifically, ViT-B/16 is adopted for both our model and CLIP2Point \cite{Huang2022CLIP}. For PointCLIP, we utilize ResNet-101~\cite{he2016deep}, ResNet-50$\times 4$~\cite{radford2021learning}, and ViT-B/16, respectively for ModelNet10, ModelNet40, and ScanObjectNN datasets, which is to fully achieve its best performance.

\vspace{-0.2cm}
\paragraph{Main Results.} In Table~\ref{table:zero_shot_classification}, we compare the zero-shot classification performance with existing approaches.
Some models require extra pre-training on 3D point cloud datasets. CLIP2Point trains a depth map encoder on ShapeNet dataset \cite{chang2015shapenet}, and then uses it for a 3D zero-shot classification task. Cheraghian \etal \cite{cheraghian2022zero} directly extracts point cloud features with a 3D encoder. They sample `seen' categories in the dataset to pre-train the model, and validate on the `unseen' categories. In contrast, PointCLIP and our V2 discard any 3D training and can directly test on 3D datasets.
For all three benchmarks, our approach outperforms existing works by significant margins. V2 achieves $73.13\%$ and $64.22\%$ accuracy on ModelNet10 and ModelNet40, respectively, surpassing PointCLIP by $+42.90\%$ and $+40.44\%$. V2 also achieves $35.36\%$ on PB\_T50\_RS split of the ScanObjectNN dataset, demonstrating our effectiveness under noisy real-world scenes.

\begin{table}[t!]
\centering
\begin{adjustbox}{width=0.91\linewidth}
	\begin{tabular}{ccccc}
	\toprule
		Caption & Question & Paraphrase & Words & Zero-shot \\ \cmidrule(lr){1-5}
		- & - & - & - & $39.11$\\
		\checkmark & - & - & - & $61.67$\\
		\checkmark & \checkmark & - & - & $60.86$\\
		\checkmark & - & \checkmark & - & $61.12$\\
		\checkmark & \checkmark & \checkmark & - & $63.29$\\
		\checkmark & \checkmark & - & \checkmark & $61.26$\\ 
		\checkmark & \checkmark & \checkmark & \checkmark & $64.22$ \\ 
	\bottomrule
	\end{tabular}
\end{adjustbox}
\vspace{0.2cm}
\caption{\textbf{Ablation Study of GPT Prompting} on ModelNet40~\cite{wu20153d} zero-shot classification (\%). We compare four types of language command to generate the 3D-specific text.}
\vspace*{-5pt}
\label{table:ablation_commands}
\end{table}

\begin{table*}[t!]
\centering
\vspace*{-0.3pt}
\begin{adjustbox}{width=\linewidth}
	\begin{tabular}{l|c|cccccccccccc}
	\toprule
		& mIoU$_I$ & Airplane & Bag & Cap & Chair & Earphone & Guitar & Knife & Laptop & Mug & Rocket & Skate & Table  \\
        \cmidrule(lr){1-14} 
        \# Shapes   &  $2874$  &$341$ &$14$  &$11$ & $704$ & $14$ & $159$ & $80$ & $83$ & $38$ & $12$ & $31$ & $848$ \\
        \cmidrule(lr){1-14} 
        PointCLIP$^*$ & $31.0$ & $22.0$ & $44.8$ & $13.4$ & $18.7$ & $28.3$ & $22.7$ & $24.8$ & $22.9$ & $48.6$ & $22.7$ & $42.7$ & $45.5$ \\
        \textbf{PointCLIP V2} & $\mathbf{49.5}$ & $\mathbf{33.5}$ & $\mathbf{60.4}$ & $\mathbf{52.8}$ & $\mathbf{51.5}$ & $\mathbf{56.5}$ & $\mathbf{71.5}$ & $\mathbf{66.7}$ & $\mathbf{61.6}$ & $\mathbf{48.0}$ & $\mathbf{49.6}$ & $\mathbf{43.9}$ & $\mathbf{61.1}$ \\
        \specialrule{0em}{1pt}{1pt}
	\bottomrule
	\end{tabular}
\end{adjustbox}
\vspace{0.15cm}
\caption{\textbf{Zero-shot Part Segmentation (\%) on ShapeNetPart~\cite{yi2016scalable}.} We implement PointCLIP by our proposed segmentation pipeline.}
\label{table:segment_iou}
\end{table*}
\begin{table*}[t!]
\centering
\begin{adjustbox}{width=\linewidth}
	\begin{tabular}{c|l|c|cccccccccccc}
	\toprule
		 & Method & Mean & Cabinet & Bed & Chair & Sofa & Table & Door & Window & Counter & Desk & Sink & Bathtub \\
        \cmidrule(lr){1-14} 
        \multirow{2}{*}{AP$_{25}$} & PointCLIP* & $6.00$ & $3.99$ & $4.82$ & $45.16$ & $4.82$ & $7.36$ & $4.62$ & $2.19$ & $1.02$ & $4.00$ & $13.40$ & $6.46$ \\
        & \textbf{PointCLIP V2} & $\mathbf{18.97}$ & $\mathbf{19.32}$ & $\mathbf{20.98}$ & $\mathbf{61.89}$ & $\mathbf{15.55}$ & $\mathbf{23.78}$ & $\mathbf{13.22}$ & $\mathbf{17.42}$ & $\mathbf{12.43}$ & $\mathbf{21.43}$ & $\mathbf{14.54}$ & $\mathbf{16.77}$ \\
        \cmidrule(lr){1-14} 
        \multirow{2}{*}{AP$_{50}$} & PointCLIP$^*$ & $4.76$ & $1.67$ & $4.33$ & $39.53$ & $3.65$ & $5.97$ & $2.61$ & $0.52$ & $0.42$ & $2.45$ & $5.27$ & $1.31$\\
        & \textbf{PointCLIP V2} & $\mathbf{11.53}$ & $\mathbf{10.43}$ & $\mathbf{13.54}$ & $\mathbf{41.23}$ & $\mathbf{6.60}$ & $\mathbf{15.21}$ & $\mathbf{6.23}$ & $\mathbf{11.35}$ & $\mathbf{6.23}$ & $\mathbf{10.84}$ & $\mathbf{11.43}$ & $\mathbf{10.14}$ \\
        \specialrule{0em}{1pt}{1pt}
	\bottomrule
	\end{tabular}
\end{adjustbox}
\vspace{0.15cm}
\caption{\textbf{Zero-shot 3D Object Detection (\%) on ScanNet V2~\cite{dai2017scannet}.} We implement PointCLIP by our proposed detection pipeline.}
\label{table:zeroshot_detection}
\end{table*}

\vspace{-0.2cm}
\paragraph{Ablation Study.}
In Table \ref{table:shape_projection}, we conduct an ablation study of PointCLIP V2 concerning four steps of the realistic projection module. When we directly project the point cloud into 2D images via orthogonal projection, the zero-shot accuracy performs $57.35\%$, reduced by $-6.87\%$. If the quantizing step is adopted, the densifying and smoothing operation can improve zero-shot performance by $+15.14\%$ and $+5.7\%$, respectively, indicating the importance of these two steps. We also compare alternative pooling operations for the densifying step, including maximum, minimum, and average pooling. We observe that the minimum pooling achieves the best performance, which is consistent with the occlusion effect in the real world. In Table \ref{table:ablation_commands}, we show the effect of four command types in the GPT prompting module. Under different command combinations, the zero-shot performance is improved with various degrees. If using all four types, the 3D-specific text improves the zero-shot performance by $+25.11\%$, indicating the great significance of better language-image alignment.

\subsection{Few-shot Classification}

\paragraph{Settings.} We test $k$-shot classification performance on ModelNet40~\cite{wu20153d} and ScanObjectNN~\cite{uy2019revisiting} datasets, where $k \in \left \{1,2,4,8,16 \right \}$. We adopt the same 3D-specific text used in the zero-shot task as textual input and jointly train the learnable smoothing (Figure \ref{fig:shape_projection}) and inter-view adapter~\cite{zhang2022pointclip}.
 The 3D convolution layers adopt a $5\times5\times3$ kernel size and are followed by a batch normalization \cite{ioffe2015batch} with a ReLU non-linear activation \cite{nair2010rectified}.

\begin{figure}[t!]
\centering
\includegraphics[width=0.47\textwidth]{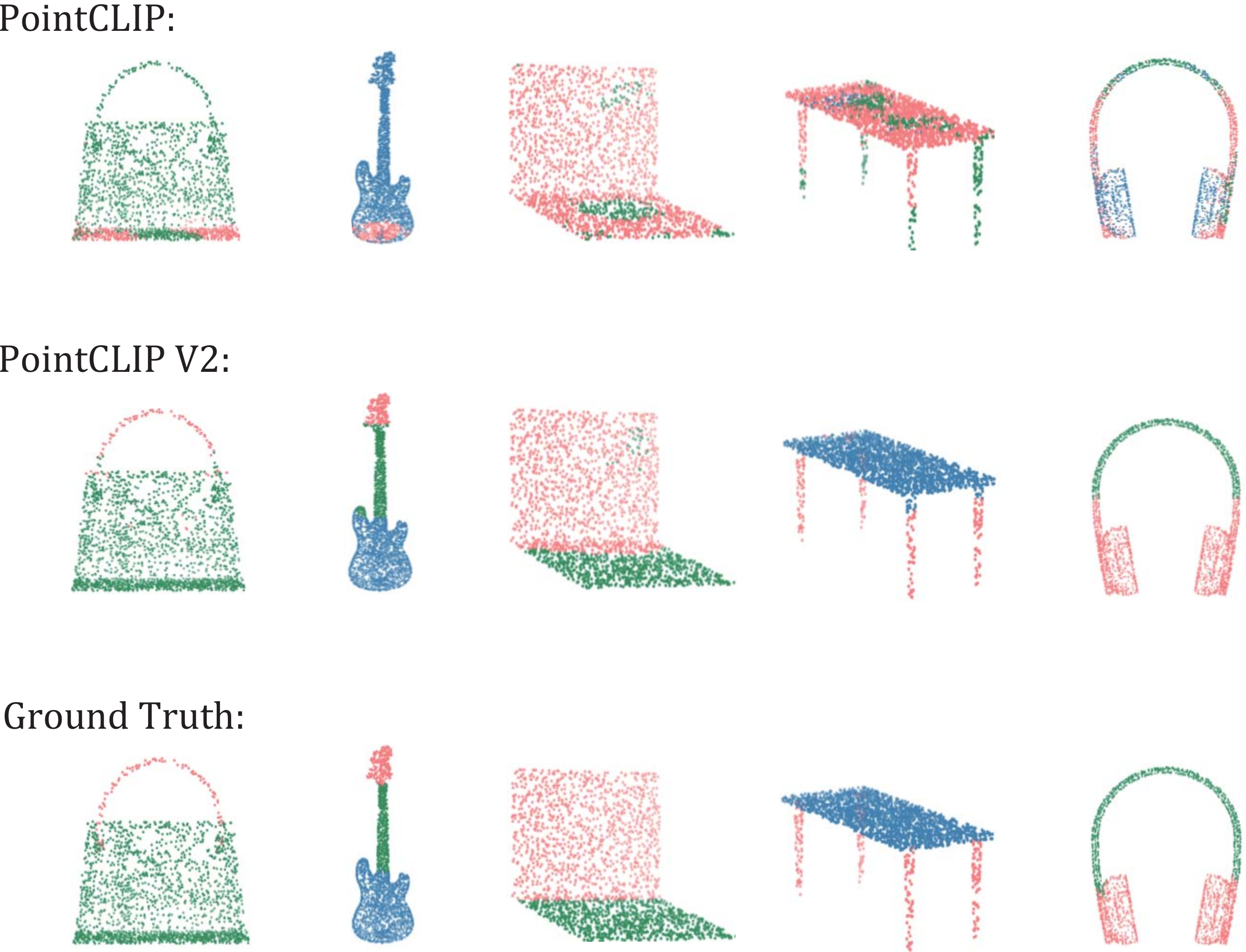}
\vspace{0.1cm}
\caption{\textbf{Visualization of Zero-shot Part Segmentation on ShapeNetPart \cite{yi2016scalable}.} Our V2 exhibits better fine-grained segmentation than PointCLIP.}
\label{fig:segment_example}
\end{figure}

\vspace{-0.1cm}
\paragraph{Main Results.} 
In Figure~\ref{fig:few_shot_classification}, we show the few-shot classification results of V2, comparing with PointCLIP and other four representative 3D networks: PointNet~\cite {qi2016pointnet}, PointNet++~\cite{qi2017pointnet++}, SimpleView~\cite{goyal2021revisiting}, and CurveNet~\cite{Xiang2021Walk}. As shown, V2 outperforms all other methods by few-shot training and shows a more significant improvement on $1$-shot classification. V2 surpasses PointCLIP's $1$-shot accuracy by $+12$\% on ModelNet40 and $+7$\% on ScanObjectNN. In addition, our approach achieves a $16$-shot accuracy of $89.55\%$ on ModelNet40 dataset, even approaching the fully supervised PointNet\cite{qi2016pointnet}.

\vspace{-0.1cm}
\paragraph{Ablation Study. }
In Table~\ref{table:few_ablation}, we report the impact of different modules on few-shot V2 with $16$-shot results, including the learnable smoothing, the view weighing following PointCLIP, and 3D-specific text from GPT prompting.
We find that the learnable 3D projection module improves $16$-shot accuracy by $+0.7\%$ than the fixed one, and adopting 3D-specific text improves accuracy by $+3.33\%$. 

\subsection{Zero-shot Part Segmentation}

\paragraph{Settings.} 
We evaluate the zero-shot segmentation performance on the ShapeNetPart dataset \cite{yi2016scalable}, which includes $16$ categories and $50$ annotated parts. Following prior methods \cite{qi2017pointnet++,wang2019dynamic,ma2022rethinking}, we sample 2048 points from each point cloud, and test on the official test split. \textit{For comparison, we implement PointCLIP via our proposed zero-shot segmentation pipeline and report the best-performing results.} 

\vspace{-0.1cm}
\paragraph{Main Results.}
We show the mean intersection of union score across instances (mIoU$_I$) in Table \ref{table:segment_iou}. Our method surpasses PointCLIP by $+17.4\%$ for overall mIoU$_I$ and performs consistently better on different object categories. We also visualize the segmentation results in Figure~\ref{fig:segment_example}, which further demonstrates our effectiveness to capture fine-grained 3D patterns in a zero-shot manner.

\subsection{Zero-shot 3D Object Detection.} 

\paragraph{Settings.}
ScanNet V2 dataset~\cite{dai2017scannet} is utilized to evaluate the detection performance, which contains 18 object categories. We adopt the pre-trained 3DETR-m~\cite{misra2021end} model as the region proposal network and extract 1024 points within each 3D box. We report the zero-shot detection performance on the validation set using mean Average Precision (mAP) at two different IoU thresholds of $0.25$ and $0.5$, denoted as AP$_{25}$ and AP$_{50}$. \textit{Also, PointCLIP is implemented by our efforts for zero-shot 3D detection and we report the best-performing results.}

\vspace{-0.1cm}
\paragraph{Main Results.}
Table \ref{table:zeroshot_detection} shows our zero-shot 3D detection results compared with PointCLIP. We observe that PointCLIP V2 achieves mAP$_{25}$ and mAP$_{50}$ of $18.97\%$ and $11.53\%$, outperforming PointCLIP by $+12.97\%$ and $+6.77\%$, respectively. This verifies that V2 is superior to recognize 3D open-world objects in real-world scenes and obtains great potential for general 3D open-world learning.

\vspace{0.1cm}

\subsection{Other Experiments}

\paragraph{Computation Burden.} We have compared the latency of inference in Table~\ref{table:ablation_projection}. Additionally, we compare the computation complexity to PointCLIP~\cite{zhang2022pointclip} and CLIP2Point~\cite{Huang2022CLIP} in Table \ref{tab:computation_burden}. We test the computation overhead of each inference on 1 RTX A6000 with ViT-B/32 backbone. From the table, V2 causes a similar overhead to PointCLIP and achieves superior zero-shot accuracy on ModelNet40. Thus we achieve a better accuracy-efficiency trade-off.

\begin{table}[t!]
\centering
\begin{adjustbox}{width=0.93\linewidth}
	\begin{tabular}{c c c c }
	\toprule
	\multirow{1}{*}{Method} &\multicolumn{1}{c}{PointCLIP} &\multicolumn{1}{c}{CLIP2Point} &\multicolumn{1}{c}{PointCLIP V2}\\
	\cmidrule(lr){1-1} \cmidrule(lr){2-2} \cmidrule(lr){3-3} \cmidrule(lr){4-4}
	GFLOPs & $16.46$ & $16.46$ & $16.51$\\
        Memory (GB) & $2901$ & $3006$ & $2967$\\
        Accuracy (\%) & $16.94$ & $49.38$ & $55.92$\\
	\bottomrule
	\end{tabular}
\end{adjustbox}
\vspace{0.3cm}
\caption{\textbf{Comparison of Accuracy and Computation Overhead} with other approaches on ModelNet40~\cite{wu20153d}. }
\label{tab:computation_burden}
\end{table}

\paragraph{More Ablations for Zero-shot Classification.}
\label{abzero}
In Table~\ref{table:zero_class_backbone} and \ref{table:zero_class_numpoint}, we additionally investigate 2 factors that influence the zero-shot classification performance: the visual encoder backbone and the number of sampled points. 
\textbf{1) Different Backbones.}
In Table \ref{table:zero_class_backbone}, we examine the results with different backbones on ModelNet40~\cite{wu20153d} and ScanObjectNN~\cite{uy2019revisiting} datasets. We observe that the default ViT-B/16 backbone achieves the best overall performance. 
\textbf{2) Sample Rate of Points.}
Table \ref{table:zero_class_numpoint} presents the effect of different numbers of sampled points. Note that the officially released ModelNet40 dataset contains only $2048$ points per point cloud, so we adopt a resampled version of ModelNet40 from \cite{wang2022p2p}, which contains $8192$ points per point cloud. We observe improvements when increasing the sampling rate of points.

\section{Conclusion}
\label{sec:conclusion}

We propose PointCLIP V2, a powerful and unified 3D open-world learner, which surpasses the existing PointCLIP with significant margins. We propose to prompt CLIP with a realistic projection module for producing high-quality depth maps from 3D, and prompt GPT-3 model to generate 3D-specific descriptions. The visual and language representations achieve better alignment via prompting. Besides classification, V2 can generalize to various challenging tasks with promising performance, including 3D few-shot classification, 3D zero-shot part segmentation, and object detection. For future work, we will further explore how to adapt CLIP to wider open-world applications, \eg, outdoor 3D detection and visual grounding.

\begin{table}[t]
\centering
\begin{adjustbox}{width=\linewidth}
	\begin{tabular}{ccccccc}
	\toprule
		Datasets &RN50 & RN101 & ViT-B/32 & ViT-B/16 & RN.$\times$4\\
        \cmidrule(lr){1-6}
        \specialrule{0em}{1pt}{1pt}
		ModelNet40 &$46.45$ & $49.34$ & $60.00$  &$\mathbf{64.22}$ & $56.28$ \\
        ScanObjectNN &$33.21$ &$31.47$ &$\mathbf{35.36}$  &$34.91$ &$34.98$ \\ 
		 \specialrule{0em}{1pt}{1pt}
	\bottomrule
	\end{tabular}
\end{adjustbox}
\vspace{1pt}
\caption{\textbf{Ablation Study on Visual Encoders} for Zero-shot Classification (\%) on ModelNet40~\cite{wu20153d} and ScanObjectNN~\cite{uy2019revisiting}.}
\label{table:zero_class_backbone}
\end{table}

\begin{table}[t!]
\centering
\vspace{0.1cm}
\begin{adjustbox}{width=0.89\linewidth}
	\begin{tabular}{ccccccc}
	\toprule
		Point Number &1024 & 2048 &3072 & 4096 & 8192 \\
        \cmidrule(lr){1-6}
        \specialrule{0em}{1pt}{1pt}
        
		 ModelNet40 &64.22 & 65.28 & 66.17 & 66.45  & \textbf{68.56}\\ 
		 \specialrule{0em}{1pt}{1pt}
        ScanObjectNN &34.91 &36.05  &36.26  &37.27 & \textbf{38.90} \\ 
		 \specialrule{0em}{1pt}{1pt}
	\bottomrule
	\end{tabular}
\end{adjustbox}
\vspace{0.3cm}
\caption{\textbf{Ablation Study on Point Number} for Zero-shot Classification (\%).}
\label{table:zero_class_numpoint}
\end{table}

\paragraph{Acknowledgement.}
This work is partially supported by the National Natural Science Foundation of China (Grant No.62206272), and by the National Key R\&D Program of China (NO.2022ZD0160100).

{\small
\bibliographystyle{ieee_fullname}
\bibliography{main}

\begin{thebibliography}{10}\itemsep=-1pt

\bibitem{ahmed2019epn}
Syeda~Mariam Ahmed, Pan Liang, and Chee~Meng Chew.
\newblock {EPN: Edge-aware PointNet for object recognition from multi-view 2.5D
  point clouds}.
\newblock In {\em IEEE International Conference on Intelligent Robots and
  Systems}, pages 3445--3450, 2019.

\bibitem{bahng2022visual}
Hyojin Bahng, Ali Jahanian, Swami Sankaranarayanan, and Phillip Isola.
\newblock Visual prompting: Modifying pixel space to adapt pre-trained models.
\newblock {\em arXiv preprint arXiv:2203.17274}, 2022.

\bibitem{brown2020language}
Tom Brown, Benjamin Mann, Nick Ryder, Melanie Subbiah, Jared~D Kaplan, Prafulla
  Dhariwal, Arvind Neelakantan, Pranav Shyam, Girish Sastry, Amanda Askell,
  et~al.
\newblock Language models are few-shot learners.
\newblock {\em Advances in Neural Information Processing Systems},
  33:1877--1901, 2020.

\bibitem{chang2015shapenet}
Angel~X Chang, Thomas Funkhouser, Leonidas Guibas, Pat Hanrahan, Qixing Huang,
  Zimo Li, Silvio Savarese, Manolis Savva, Shuran Song, Hao Su, et~al.
\newblock {ShapeNet}: An information-rich {3D} model repository.
\newblock {\em arXiv preprint arXiv:1512.03012}, 2015.

\bibitem{chen2022visual}
Aochuan Chen, Peter Lorenz, Yuguang Yao, Pin-Yu Chen, and Sijia Liu.
\newblock Visual prompting for adversarial robustness.
\newblock {\em arXiv preprint arXiv:2210.06284}, 2022.

\bibitem{cheraghian2019mitigating}
Ali Cheraghian, Shafin Rahman, Dylan Campbell, and Lars Petersson.
\newblock Mitigating the hubness problem for zero-shot learning of {3D}
  objects.
\newblock {\em arXiv preprint arXiv:1907.06371}, 2019.

\bibitem{cheraghian2022zero}
Ali Cheraghian, Shafin Rahman, Townim~F Chowdhury, Dylan Campbell, and Lars
  Petersson.
\newblock Zero-shot learning on 3d point cloud objects and beyond.
\newblock {\em International Journal of Computer Vision}, pages 1--21, 2022.

\bibitem{cheraghian2019zero}
Ali Cheraghian, Shafin Rahman, and Lars Petersson.
\newblock Zero-shot learning of {3D} point cloud objects.
\newblock In {\em IEEE International Conference on Machine Vision
  Applications}, pages 1--6, 2019.

\bibitem{dai2017scannet}
Angela Dai, Angel~X Chang, Manolis Savva, Maciej Halber, Thomas Funkhouser, and
  Matthias Nie{\ss}ner.
\newblock {ScanNet}: Richly-annotated {3D} reconstructions of indoor scenes.
\newblock In {\em IEEE Conference on Computer Vision and Pattern Recognition},
  pages 5828--5839, 2017.

\bibitem{dosovitskiy2020image}
Alexey Dosovitskiy, Lucas Beyer, Alexander Kolesnikov, Dirk Weissenborn,
  Xiaohua Zhai, Thomas Unterthiner, Mostafa Dehghani, Matthias Minderer, Georg
  Heigold, Sylvain Gelly, et~al.
\newblock An image is worth 16x16 words: Transformers for image recognition at
  scale.
\newblock {\em arXiv preprint arXiv:2010.11929}, 2020.

\bibitem{fu2022pos}
Kexue Fu, Peng Gao, ShaoLei Liu, Renrui Zhang, Yu Qiao, and Manning Wang.
\newblock Pos-bert: Point cloud one-stage bert pre-training.
\newblock {\em arXiv preprint arXiv:2204.00989}, 2022.

\bibitem{gan2023decorate}
Yulu Gan, Yan Bai, Yihang Lou, Xianzheng Ma, Renrui Zhang, Nian Shi, and Lin
  Luo.
\newblock Decorate the newcomers: Visual domain prompt for continual test time
  adaptation.
\newblock {\em arXiv preprint arXiv:2212.04145}, 2023.

\bibitem{gao2021clip}
Peng Gao, Shijie Geng, Renrui Zhang, Teli Ma, Rongyao Fang, Yongfeng Zhang,
  Hongsheng Li, and Yu Qiao.
\newblock {CLIP-Adapter}: Better vision-language models with feature adapters.
\newblock {\em arXiv preprint arXiv:2110.04544}, 2021.

\bibitem{goyal2021revisiting}
Ankit Goyal, Hei Law, Bowei Liu, Alejandro Newell, and Jia Deng.
\newblock Revisiting point cloud shape classification with a simple and
  effective baseline.
\newblock {\em International Conference on Machine Learning}, 2021.

\bibitem{gu2021open}
Xiuye Gu, Tsung-Yi Lin, Weicheng Kuo, and Yin Cui.
\newblock Open-vocabulary object detection via vision and language knowledge
  distillation.
\newblock {\em arXiv preprint arXiv:2104.13921}, 2021.

\bibitem{guo2023viewrefer}
Ziyu Guo, Yiwen Tang, Renrui Zhang, Dong Wang, Zhigang Wang, Bin Zhao, and
  Xuelong Li.
\newblock Viewrefer: Grasp the multi-view knowledge for 3d visual grounding
  with gpt and prototype guidance.
\newblock {\em arXiv preprint arXiv:2303.16894}, 2023.

\bibitem{guo2023joint}
Ziyu Guo, Renrui Zhang, Longtian Qiu, Xianzhi Li, and Pheng~Ann Heng.
\newblock Joint-mae: 2d-3d joint masked autoencoders for 3d point cloud
  pre-training.
\newblock {\em IJCAI 2023}, 2023.

\bibitem{he2016deep}
Kaiming He, Xiangyu Zhang, Shaoqing Ren, and Jian Sun.
\newblock Deep residual learning for image recognition.
\newblock In {\em Proceedings of the IEEE conference on computer vision and
  pattern recognition}, pages 770--778, 2016.

\bibitem{Huang2022CLIP}
Tianyu Huang, Bowen Dong, Yunhan Yang, Xiaoshui Huang, Rynson~WH Lau, Wanli
  Ouyang, and Wangmeng Zuo.
\newblock {CLIP2Point}: Transfer {CLIP} to point cloud classification with
  image-depth pre-training.
\newblock {\em arXiv preprint arXiv:2210.01055}, 2022.

\bibitem{ioffe2015batch}
Sergey Ioffe and Christian Szegedy.
\newblock Batch normalization: Accelerating deep network training by reducing
  internal covariate shift.
\newblock In {\em International Conference on Machine Learning}, pages
  448--456, 2015.

\bibitem{jia2022visual}
Menglin Jia, Luming Tang, Bor-Chun Chen, Claire Cardie, Serge Belongie, Bharath
  Hariharan, and Ser-Nam Lim.
\newblock Visual prompt tuning.
\newblock In {\em European Conference on Computer Vision}, pages 709--727,
  2022.

\bibitem{Jiang2020How}
Zhengbao Jiang, Frank~F. Xu, Jun Araki, and Graham Neubig.
\newblock How can we know what language models know?
\newblock {\em Transactions of the Association for Computational Linguistics},
  8:423--438, 2020.

\bibitem{lin2022frozen}
Ziyi Lin, Shijie Geng, Renrui Zhang, Peng Gao, Gerard de Melo, Xiaogang Wang,
  Jifeng Dai, Yu Qiao, and Hongsheng Li.
\newblock Frozen clip models are efficient video learners.
\newblock In {\em Computer Vision--ECCV 2022: 17th European Conference, Tel
  Aviv, Israel, October 23--27, 2022, Proceedings, Part XXXV}, pages 388--404.
  Springer, 2022.

\bibitem{liu2021segmenting}
Bo Liu, Shuang Deng, Qiulei Dong, and Zhanyi Hu.
\newblock Segmenting {3D} hybrid scenes via zero-shot learning.
\newblock {\em arXiv preprint arXiv:2107.00430}, 2021.

\bibitem{liu2021pre}
Pengfei Liu, Weizhe Yuan, Jinlan Fu, Zhengbao Jiang, Hiroaki Hayashi, and
  Graham Neubig.
\newblock Pre-train, prompt, and predict: A systematic survey of prompting
  methods in natural language processing.
\newblock {\em arXiv preprint arXiv:2107.13586}, 2021.

\bibitem{liu2019roberta}
Yinhan Liu, Myle Ott, Naman Goyal, Jingfei Du, Mandar Joshi, Danqi Chen, Omer
  Levy, Mike Lewis, Luke Zettlemoyer, and Veselin Stoyanov.
\newblock Roberta: A robustly optimized bert pretraining approach.
\newblock {\em arXiv preprint arXiv:1907.11692}, 2019.

\bibitem{Lu2022Open}
Yuheng Lu, Chenfeng Xu, Xiaobao Wei, Xiaodong Xie, Masayoshi Tomizuka, Kurt
  Keutzer, and Shanghang Zhang.
\newblock Open-vocabulary {3D} detection via image-level class and debiased
  cross-modal contrastive learning.
\newblock 2022.

\bibitem{ma2022rethinking}
Xu Ma, Can Qin, Haoxuan You, Haoxi Ran, and Yun Fu.
\newblock Rethinking network design and local geometry in point cloud: A simple
  residual {MLP} framework.
\newblock {\em arXiv preprint arXiv:2202.07123}, 2022.

\bibitem{meng2021towards}
Qinghao Meng, Wenguan Wang, Tianfei Zhou, Jianbing Shen, Yunde Jia, and Luc
  Van~Gool.
\newblock Towards a weakly supervised framework for {3D} point cloud object
  detection and annotation.
\newblock {\em IEEE Transactions on Pattern Analysis and Machine Intelligence},
  2021.

\bibitem{michele2021generative}
Bj{\"o}rn Michele, Alexandre Boulch, Gilles Puy, Maxime Bucher, and Renaud
  Marlet.
\newblock Generative zero-shot learning for semantic segmentation of {3D} point
  clouds.
\newblock In {\em IEEE International Conference on 3D Vision}, pages 992--1002,
  2021.

\bibitem{misra2021end}
Ishan Misra, Rohit Girdhar, and Armand Joulin.
\newblock An end-to-end transformer model for 3d object detection.
\newblock In {\em International Conference on Computer Vision}, pages
  2906--2917, 2021.

\bibitem{naeem20223d}
Muhammad~Ferjad Naeem, Evin~P{\i}nar {\"O}rnek, Yongqin Xian, Luc Van~Gool, and
  Federico Tombari.
\newblock {3D} compositional zero-shot learning with decompositional consensus.
\newblock In {\em European Conference on Computer Vision}, pages 713--730,
  2022.

\bibitem{nair2010rectified}
Vinod Nair and Geoffrey~E Hinton.
\newblock Rectified linear units improve {Restricted Boltzmann} machines.
\newblock In {\em International Conference on Machine Learning}, 2010.

\bibitem{pratt2022Whatdoes}
Sarah Pratt, Rosanne Liu, and Ali Farhadi.
\newblock What does a platypus look like? {Generating} customized prompts for
  zero-shot image classification.
\newblock {\em arXiv preprint arXiv:2209.03320}, 2022.

\bibitem{qi2016pointnet}
Charles~R Qi, Hao Su, Kaichun Mo, and Leonidas~J Guibas.
\newblock {PointNet}: Deep learning on point sets for {3D} classification and
  segmentation.
\newblock {\em arXiv preprint arXiv:1612.00593}, 2016.

\bibitem{qi2017pointnet++}
Charles~Ruizhongtai Qi, Li Yi, Hao Su, and Leonidas~J Guibas.
\newblock {PointNet++}: Deep hierarchical feature learning on point sets in a
  metric space.
\newblock {\em Advances in Neural Information Processing Systems}, 30, 2017.

\bibitem{radford2021learning}
Alec Radford, Jong~Wook Kim, Chris Hallacy, Aditya Ramesh, Gabriel Goh,
  Sandhini Agarwal, Girish Sastry, Amanda Askell, Pamela Mishkin, Jack Clark,
  Gretchen Krueger, and Ilya Sutskever.
\newblock Learning transferable visual models from natural language
  supervision, 2021.

\bibitem{radford2018improving}
Alec Radford, Karthik Narasimhan, Tim Salimans, Ilya Sutskever, et~al.
\newblock Improving language understanding by generative pre-training.
\newblock 2018.

\bibitem{radford2019language}
Alec Radford, Jeffrey Wu, Rewon Child, David Luan, Dario Amodei, Ilya
  Sutskever, et~al.
\newblock Language models are unsupervised multitask learners.
\newblock {\em OpenAI blog}, 1(8):9, 2019.

\bibitem{raffel2020exploring}
Colin Raffel, Noam Shazeer, Adam Roberts, Katherine Lee, Sharan Narang, Michael
  Matena, Yanqi Zhou, Wei Li, and Peter~J Liu.
\newblock Exploring the limits of transfer learning with a unified text-to-text
  transformer.
\newblock {\em The Journal of Machine Learning Research}, 21(1):5485--5551,
  2020.

\bibitem{Shi2015DeepPano}
Baoguang Shi, Song Bai, Zhichao Zhou, and Xiang Bai.
\newblock {DeepPano}: Deep panoramic representation for {3-D} shape
  recognition.
\newblock {\em IEEE Signal Processing Letters}, 22(12):2339--2343, 2015.

\bibitem{su2015multi}
Hang Su, Subhransu Maji, Evangelos Kalogerakis, and Erik Learned-Miller.
\newblock Multi-view convolutional neural networks for {3D} shape recognition.
\newblock In {\em International Conference on Computer Vision}, pages 945--953,
  2015.

\bibitem{su2018adeeper}
Jong-Chyi Su, Matheus Gadelha, Rui Wang, and Subhransu Maji.
\newblock A deeper look at {3D} shape classifiers.
\newblock In {\em European Conference on Computer Vision Workshops}, 2018.

\bibitem{uy2019revisiting}
Mikaela~Angelina Uy, Quang-Hieu Pham, Binh-Son Hua, Thanh Nguyen, and Sai-Kit
  Yeung.
\newblock Revisiting point cloud classification: A new benchmark dataset and
  classification model on real-world data.
\newblock In {\em International Conference on Computer Vision}, pages
  1588--1597, 2019.

\bibitem{vaswani2017attention}
Ashish Vaswani, Noam Shazeer, Niki Parmar, Jakob Uszkoreit, Llion Jones,
  Aidan~N Gomez, {\L}ukasz Kaiser, and Illia Polosukhin.
\newblock Attention is all you need.
\newblock {\em Advances in Neural Information Processing Systems}, 30, 2017.

\bibitem{wallace2019universal}
Eric Wallace, Shi Feng, Nikhil Kandpal, Matt Gardner, and Sameer Singh.
\newblock Universal adversarial triggers for attacking and analyzing {NLP}.
\newblock In {\em Conference on Empirical Methods in Natural Language
  Processing and the 9th International Joint Conference on Natural Language
  Processing (EMNLP-IJCNLP)}, pages 2153--2162, Nov. 2019.

\bibitem{wang2019dominant}
Chu Wang, Marcello Pelillo, and Kaleem Siddiqi.
\newblock Dominant set clustering and pooling for multi-view {3D} object
  recognition.
\newblock {\em arXiv preprint arXiv:1906.01592}, 2019.

\bibitem{wang2019dynamic}
Yue Wang, Yongbin Sun, Ziwei Liu, Sanjay~E Sarma, Michael~M Bronstein, and
  Justin~M Solomon.
\newblock Dynamic graph {CNN} for learning on point clouds.
\newblock {\em ACM Transactions On Graphics}, 38(5):1--12, 2019.

\bibitem{wang2022p2p}
Ziyi Wang, Xumin Yu, Yongming Rao, Jie Zhou, and Jiwen Lu.
\newblock {P2P}: Tuning pre-trained image models for point cloud analysis with
  point-to-pixel prompting.
\newblock {\em arXiv preprint arXiv:2208.02812}, 2022.

\bibitem{wei2020view}
Xin Wei, Ruixuan Yu, and Jian Sun.
\newblock {View-GCN}: View-based graph convolutional network for {3D} shape
  analysis.
\newblock In {\em IEEE Conference on Computer Vision and Pattern Recognition},
  pages 1850--1859, 2020.

\bibitem{wu2019pointconv}
Wenxuan Wu, Zhongang Qi, and Li Fuxin.
\newblock {PointConv}: Deep convolutional networks on {3D} point clouds.
\newblock In {\em IEEE Conference on Computer Vision and Pattern Recognition},
  pages 9621--9630, 2019.

\bibitem{wu2022eda}
Yanmin Wu, Xinhua Cheng, Renrui Zhang, Zesen Cheng, and Jian Zhang.
\newblock Eda: Explicit text-decoupling and dense alignment for 3d visual and
  language learning.
\newblock {\em arXiv preprint arXiv:2209.14941}, 2022.

\bibitem{wu20153d}
Zhirong Wu, Shuran Song, Aditya Khosla, Fisher Yu, Linguang Zhang, Xiaoou Tang,
  and Jianxiong Xiao.
\newblock {3D ShapeNets}: A deep representation for volumetric shapes.
\newblock In {\em IEEE Conference on Computer Vision and Pattern Recognition},
  pages 1912--1920, 2015.

\bibitem{Xiang2021Walk}
Tiange Xiang, Chaoyi Zhang, Yang Song, Jianhui Yu, and Weidong Cai.
\newblock Walk in the cloud: Learning curves for point clouds shape analysis.
\newblock In {\em International Conference on Computer Vision}, pages 915--924,
  2021.

\bibitem{xu2020squeezesegv3}
Chenfeng Xu, Bichen Wu, Zining Wang, Wei Zhan, Peter Vajda, Kurt Keutzer, and
  Masayoshi Tomizuka.
\newblock {SqueezeSegV3}: Spatially-adaptive convolution for efficient
  point-cloud segmentation.
\newblock In {\em European Conference on Computer Vision}, pages 1--19, 2020.

\bibitem{yang2019xlnet}
Zhilin Yang, Zihang Dai, Yiming Yang, Jaime Carbonell, Russ~R Salakhutdinov,
  and Quoc~V Le.
\newblock Xlnet: Generalized autoregressive pretraining for language
  understanding.
\newblock {\em Advances in Neural Information Processing Systems}, 32, 2019.

\bibitem{yang2022empirical}
Zhengyuan Yang, Zhe Gan, Jianfeng Wang, Xiaowei Hu, Yumao Lu, Zicheng Liu, and
  Lijuan Wang.
\newblock An empirical study of {GPT}-3 for few-shot knowledge-based {VQA}.
\newblock In {\em AAAI Conference on Artificial Intelligence}, volume~36, pages
  3081--3089, 2022.

\bibitem{yang2019learning}
Ze Yang and Liwei Wang.
\newblock Learning relationships for multi-view {3D} object recognition.
\newblock In {\em International Conference on Computer Vision}, pages
  7505--7514, 2019.

\bibitem{yi2016scalable}
Li Yi, Vladimir~G Kim, Duygu Ceylan, I-Chao Shen, Mengyan Yan, Hao Su, Cewu Lu,
  Qixing Huang, Alla Sheffer, and Leonidas Guibas.
\newblock A scalable active framework for region annotation in {3D} shape
  collections.
\newblock {\em ACM Transactions on Graphics}, 35(6):1--12, 2016.

\bibitem{zhang2021tip}
Renrui Zhang, Rongyao Fang, Peng Gao, Wei Zhang, Kunchang Li, Jifeng Dai, Yu
  Qiao, and Hongsheng Li.
\newblock Tip-{Adapter}: Training-free {CLIP}-adapter for better
  vision-language modeling.
\newblock {\em arXiv preprint arXiv:2111.03930}, 2021.

\bibitem{zhang2022point}
Renrui Zhang, Ziyu Guo, Peng Gao, Rongyao Fang, Bin Zhao, Dong Wang, Yu Qiao,
  and Hongsheng Li.
\newblock Point-m2ae: multi-scale masked autoencoders for hierarchical point
  cloud pre-training.
\newblock {\em arXiv preprint arXiv:2205.14401}, 2022.

\bibitem{zhang2022pointclip}
Renrui Zhang, Ziyu Guo, Wei Zhang, Kunchang Li, Xupeng Miao, Bin Cui, Yu Qiao,
  Peng Gao, and Hongsheng Li.
\newblock {PointCLIP}: Point cloud understanding by {CLIP}.
\newblock In {\em IEEE Conference on Computer Vision and Pattern Recognition},
  pages 8552--8562, 2022.

\bibitem{zhang2023prompt}
Renrui Zhang, Xiangfei Hu, Bohao Li, Siyuan Huang, Hanqiu Deng, Hongsheng Li,
  Yu Qiao, and Peng Gao.
\newblock Prompt, generate, then cache: Cascade of foundation models makes
  strong few-shot learners.
\newblock {\em arXiv preprint arXiv:2303.02151}, 2023.

\bibitem{zhang2023personalize}
Renrui Zhang, Zhengkai Jiang, Ziyu Guo, Shilin Yan, Junting Pan, Hao Dong, Peng
  Gao, and Hongsheng Li.
\newblock Personalize segment anything model with one shot.
\newblock {\em arXiv preprint arXiv:2305.03048}, 2023.

\bibitem{zhang2022learning}
Renrui Zhang, Liuhui Wang, Yu Qiao, Peng Gao, and Hongsheng Li.
\newblock Learning 3d representations from 2d pre-trained models via
  image-to-point masked autoencoders.
\newblock {\em arXiv preprint arXiv:2212.06785}, 2022.

\bibitem{zhang2023parameter}
Renrui Zhang, Liuhui Wang, Yali Wang, Peng Gao, Hongsheng Li, and Jianbo Shi.
\newblock Parameter is not all you need: Starting from non-parametric networks
  for 3d point cloud analysis.
\newblock {\em arXiv preprint arXiv:2303.08134}, 2023.

\bibitem{zhong2022regionclip}
Yiwu Zhong, Jianwei Yang, Pengchuan Zhang, Chunyuan Li, Noel Codella,
  Liunian~Harold Li, Luowei Zhou, Xiyang Dai, Lu Yuan, Yin Li, et~al.
\newblock Regionclip: Region-based language-image pretraining.
\newblock In {\em Proceedings of the IEEE/CVF Conference on Computer Vision and
  Pattern Recognition}, pages 16793--16803, 2022.

\bibitem{zhou2022cocoop}
Kaiyang Zhou, Jingkang Yang, Chen~Change Loy, and Ziwei Liu.
\newblock Conditional prompt learning for vision-language models.
\newblock In {\em IEEE Conference on Computer Vision and Pattern Recognition},
  2022.

\bibitem{zhou2022coop}
Kaiyang Zhou, Jingkang Yang, Chen~Change Loy, and Ziwei Liu.
\newblock Learning to prompt for vision-language models.
\newblock {\em International Journal of Computer Vision}, 2022.

\end{thebibliography}
}

\end{document}